\documentclass{article}

\usepackage{microtype}
\usepackage{graphicx}
\usepackage{booktabs}

\usepackage{hyperref}

\usepackage[accepted]{icml2026}

\makeatletter
\renewcommand{\ICML@appearing}{\textit{Mechanistic Interpretability Workshop at the
43rd International Conference on Machine Learning},
Seoul, South Korea, 2026. Copyright 2026 by the author(s).}
\renewcommand{\Notice@String}{\ICML@appearing}
\makeatother

\usepackage{amsmath}
\usepackage{amssymb}
\usepackage{mathtools}
\usepackage{amsthm}
\usepackage{dsfont}

\usepackage{xcolor}
\usepackage{soul}
\usepackage{array}
\usepackage{subfigure}
\usepackage{caption}
\usepackage{float}
\usepackage{multirow,bigstrut}
\usepackage{adjustbox}

\usepackage{amsmath,amsfonts,bm}

\def\eqref#1{equation~\ref{#1}}

\def\1{\bm{1}}

\DeclareMathAlphabet{\mathsfit}{\encodingdefault}{\sfdefault}{m}{sl}
\SetMathAlphabet{\mathsfit}{bold}{\encodingdefault}{\sfdefault}{bx}{n}

\DeclareMathOperator*{\argmax}{arg\,max}
\DeclareMathOperator*{\argmin}{arg\,min}

\theoremstyle{plain}
\newtheorem{theorem}{Theorem}[section]

\newtheorem{conjecture}[theorem]{Conjecture}
\theoremstyle{definition}

\icmltitlerunning{On Pitfalls of RemOve-And-Retrain: A Data Processing Inequality Perspective}

\begin{document}

\twocolumn[
  \icmltitle{On Pitfalls of \emph{RemOve-And-Retrain}:\\A Data Processing Inequality Perspective}

  \begin{icmlauthorlist}
    \icmlauthor{Junhwa Song}{pitin}
    \icmlauthor{Keumgang Cha}{sia}
    \icmlauthor{Junghoon Seo}{pitin}
  \end{icmlauthorlist}

  \icmlaffiliation{pitin}{PIT-IN Corp., Anyang-si, South Korea}
  \icmlaffiliation{sia}{SI Analytics Co., Ltd., Daejeon-si, South Korea}

  \icmlcorrespondingauthor{Junghoon Seo}{sjh@pitin-ev.com}

  \icmlkeywords{Mechanistic Interpretability, Attribution Methods, Evaluation Metrics, ROAR, Data Processing Inequality, Benchmarking}

  \vskip 0.3in
]

\printAffiliationsAndNotice{}

\begin{abstract}
The RemOve-And-Retrain (ROAR) benchmark is widely used to evaluate feature attribution methods, yet its validity remains underexplored from an information-theoretic perspective.
We show that model- and data-agnostic post-processing of attribution maps (transformations that, by the data processing inequality, \emph{cannot} add information about the decision function) can often improve ROAR scores.
This means that an improved ROAR ranking is not, by itself, evidence that an attribution map carries more information about the model. We trace this failure mode to a bias toward spatially blurry masks.
Experiments on CIFAR-10, SVHN, and CUB-200 show a consistent association between blurriness and ROAR performance, a pattern that also appears in the ROAD variant.
We provide guidelines for more cautious removal-based benchmarking, with implications for validating mechanistic understanding of neural network internals.
\end{abstract}

\section{Introduction}
\label{sec:intro}

Mechanistic interpretability seeks to understand the internal computations of neural networks, from identifying task-relevant circuits~\citep{wang2023interpretability} and validating hypotheses about learned representations~\citep{bricken2023monosemanticity} to reverse-engineering decision-making processes~\citep{sharkey2025open}.
Feature attribution methods, which measure the contribution of each input feature to a model's decisions, are foundational building blocks in this enterprise.
They underpin circuit discovery, concept validation, and safety-oriented model auditing, making their \emph{reliable evaluation} a first-order concern for the field.
Without sound benchmarks, we cannot distinguish methods that genuinely reveal decision-relevant structure from those that merely produce visually appealing but uninformative saliency maps~\citep{samek2021explaining}.

Despite significant efforts to develop quantitative evaluation methods~\citep{samek2016evaluating,dabkowski2017real,adebayo2018sanity,hooker2019benchmark,adebayo2020debugging}, selecting the most effective metric remains contentious~\citep{duan2024evaluation}.
One influential method is the RemOve-And-Retrain (ROAR) protocol~\citep{hooker2019benchmark}, widely adopted for benchmarking attribution methods~\citep{schramowski2020making,10.1145/3394486.3403071,kim2019saliency,yang2020learning,o2020generative,khakzar2021neural,hartley2021swag,chefer2021transformer,ismail2021improving,meng2022interpretability,hong2025comprehensive} and for developing novel evaluation metrics~\citep{bhatt2020evaluating,ismail2020benchmarking,deyoung-etal-2020-eraser,zhang-etal-2021-sample}. Additionally, \cite{kim2019bridging} extended ROAR to measure fixed attribution interpretability of various model functions $f(\cdot;\theta)$.

In this study, we question the reliability of the ROAR metric.
As attribution evaluation directly affects which methods are trusted in downstream mechanistic analyses and safety-relevant model auditing, understanding the failure modes of widely-used benchmarks is essential.
Our key contributions are:
\begin{itemize}
  \item We prove, via the data processing inequality, that model/data-agnostic post-processing of attributions can \emph{improve} ROAR scores while carrying \emph{less} information about the decision function --- a concrete hypothesis that we test empirically.
  \item We provide a structural causal model of the ROAR data-generation process, a formal counterexample, and experiments on real-world datasets.
  \item We reveal a persistent blurriness bias in ROAR (and ROAD) metrics, and offer practical guidelines for more cautious benchmarking of interpretability methods.
\end{itemize}

\section{Preliminaries}
\subsection{Notions and Notations}
Let $\mathcal{X}$ and $\mathcal{Y}$ denote the input space and class label space, respectively. We consider a $C$-class multi-class pre-softmax classifier $f(x; \theta): \mathcal{X} \times \Theta \rightarrow \mathbb{R}^C$ with parameters $\theta$, and define $\mathcal{F}$ as the set of all possible functions $f(\cdot; \theta): \mathcal{X} \rightarrow \mathbb{R}^C$ for a given $\theta$.
A feature importance measure (or explainer), also referred to as an attribution method~\citep{ancona2018towards}, is a function $e(x, f(\cdot; \theta), y): \mathcal{X} \times \mathcal{F} \times \mathcal{Y} \rightarrow \tilde{\mathcal{X}}$ that identifies which input features are important in determining a class decision $y \in \mathcal{Y}$ given an input $x$ and a function $f(\cdot; \theta)$~\citep{NIPS2017_8a20a862,hooker2019benchmark}.
As an example, input-gradient~\citep{JMLR:v11:baehrens10a,simonyan2014deep}, a widely recognized basic feature importance measure, is defined as $e(x, f(\cdot; \theta), y) = \frac{\partial {f(\text{x}; \theta)}y}{\partial \text{x}}\Bigr\rvert_{\text{x} = x}$, where ${f(\text{x}; \theta)}_y$ is the $y$-th indexed value of $f(\text{x}; \theta)$.
In this paper, the output of the explainer is referred to as an ``attribution map.'' For consistency, variables are denoted using uppercase letters, while functions or values are denoted using lowercase letters, with some exceptions.

\begin{algorithm}[t]
\caption{\textbf{R}em\textbf{O}ve-\textbf{A}nd-\textbf{R}etrain (ROAR) with attribution post-processing}
\label{algorithm:roar}
\small
\begin{algorithmic}[1]
\REQUIRE $f(x;\theta)$: classifier; $e(x,f(\cdot;\theta),y)$: attribution; $k(a)$: post-processing; $D_{train}$, $D_{test}$; $T$: drop rates.
\ENSURE $V$: \{drop rate $\to$ ROAR accuracy\}.
\STATE $\hat{\theta} \gets \argmin_\theta \mathbb{E}_{(x,y)\in D_{train}}[L(f(x;\theta),y)]$;\; $V \gets \{\}$ \label{algo:line1}
\STATE $A_{tr} \gets [e(x,f(\cdot;\hat{\theta}),\argmax_i f_i(x;\hat{\theta}))$ for $(x,y) \in D_{train}]$ \label{algo:line2}
\STATE $A_{te} \gets [e(x,f(\cdot;\hat{\theta}),\argmax_i f_i(x;\hat{\theta}))$ for $(x,y) \in D_{test}]$ \label{algo:line3}
\STATE $A_{tr} \gets [k(a)$ for $a \in A_{tr}]$;\; $A_{te} \gets [k(a)$ for $a \in A_{te}]$ \label{algo:line4}
\FOR{$t \in T$}
\STATE $M_{tr} \gets [\{a > \text{Pctl}(a,t)\}$ for $a \in A_{tr}]$ \label{algo:line7}
\STATE $M_{te} \gets [\{a > \text{Pctl}(a,t)\}$ for $a \in A_{te}]$ \label{algo:line8}
\STATE $D'_{tr} \gets [((1\!-\!m)\!\odot\! x,\, y)$ for $((x,y),m) \in \text{zip}(D_{train},M_{tr})]$ \label{algo:line9}
\STATE $D'_{te} \gets [((1\!-\!m)\!\odot\! x,\, y)$ for $((x,y),m) \in \text{zip}(D_{test},M_{te})]$ \label{algo:line10}
\STATE $\hat{\theta}_{new} \gets \argmin_\theta \mathbb{E}_{(x',y)\in D'_{tr}}[L(f(x'),y)]$ \label{algo:line11}
\STATE $Acc \gets \mathbb{E}_{(x',y)\in D'_{te}}[\mathds{1}_{\argmax_i f_i(x';\hat{\theta}_{new})=y}]$ \label{algo:line12}
\STATE $V[t] \gets Acc$ \label{algo:line13}
\ENDFOR
\end{algorithmic}
\end{algorithm}

\subsection{\textbf{R}em\textbf{O}ve-\textbf{A}nd-\textbf{R}etrain (ROAR)}
\label{sec:roar}
To make our work self-contained, we present a brief introduction here~\citep{hooker2019benchmark}. Algorithm 1 outlines the pseudo-code for the ROAR procedure in Python style. The proposal of ROAR was motivated by the limitations of existing modification-based evaluation methods~\citep{bach2015pixel,samek2016evaluating}. At the time, the dominant approach for modification-based evaluation was a type of sequential procedure that only used the test dataset $D_{test}$. This approach involves applying attribution methods (line 3), sorting feature importance estimates and modifying features with high attribution values (lines 8 and 10), and measuring the performance drops of the trained classifier (lines 12 and 13). It should be noted that, in this case, $\hat{\theta}_{new} \gets \hat{\theta}$ because the parameters of the classifier are never changed. However, \cite{dabkowski2017real,fong2017interpretable} have pointed out that it is difficult to determine whether the performance drops are due to the significance of the feature importance estimates or to the out-of-distribution nature of the input samples~\citep{zheng2025ffidelity}.

To address the issue of a distribution gap between training data and testing data, the ROAR approach employs a re-training strategy. This involves applying data processing techniques to the training data, $D_{train}$, in a similar manner to how they are applied to the testing data in the previously described modification-based evaluation. Specifically, after sorting the attribution for each instance based on feature importance (Line \ref{algo:line9}), a new training dataset, $D^\prime_{train}$, is created by dropping a certain percentage of features (as determined by the parameter $t$, see Line \ref{algo:line7}). The model is then re-trained on this synthesized training dataset to produce new parameters, $\hat{\theta}_{new}$ (Line \ref{algo:line11}). These new parameters are used to evaluate model performance, rather than the original parameters, $\hat{\theta}$, in order to alleviate the distribution mismatch between the training and testing datasets (Line \ref{algo:line12}).

The recent work~\citep{rong2022consistent} has demonstrated that mutual information (MI) between data variable and class variable can be utilized as a surrogate for attainable accuracy in pixel removal strategies that involve retraining, as higher MI generally leads to higher accuracy~\citep{hellman1970probability,schramowski2020making}. In the ROAR approach, MI between the modified data variable and the class variable is particularly important, as it determines the obtainable accuracy. In this context, low MI between the modified data variable and the class variable is desirable, as it results in a decrease in accuracy and improved benchmarking results.

\section{Sanity Check in View of DPI}
\label{sec:bias}

\subsection{Sketch of Our Argument}
\label{sec:argument}

A central intuition behind ROAR is that, after re-training on the modified dataset, the attainable test accuracy at a given drop rate is governed by how much class-relevant information remains in the modified input.
In particular, \citet{rong2022consistent} argue (under mild regularity assumptions) that larger mutual information between the modified input and the label typically allows higher re-trained accuracy, and conversely that lower mutual information implies lower attainable accuracy.
Motivated by this, we view the ROAR score at drop rate $t$ as a proxy for $I(X_t^\prime;Y)$,
where $X_t^\prime$ denotes the random variable of the ROAR-modified input at drop rate $t$.
Hence, \emph{a lower ROAR accuracy corresponds to a smaller} $I(X_t^\prime;Y)$, which is ``better'' in the ROAR sense.

\paragraph{What ROAR is supposed to measure.}
Conceptually, ROAR is intended to rank explainers by how well their attributions reflect the model's decision mechanism:
if an explainer highlights features that the model truly relies on for predicting $Y$, then removing those features (and retraining) should destroy predictive information and reduce accuracy.
The potential pitfall we stress is that ROAR \emph{only observes the modified data} $X^\prime$ (equivalently, the mask), and thus it is not immediate that a low $I(X^\prime;Y)$ must be caused by an attribution that is informative about the model/decision function.

\begin{figure}[t]
  \centering
  \includegraphics[width=0.45\columnwidth]{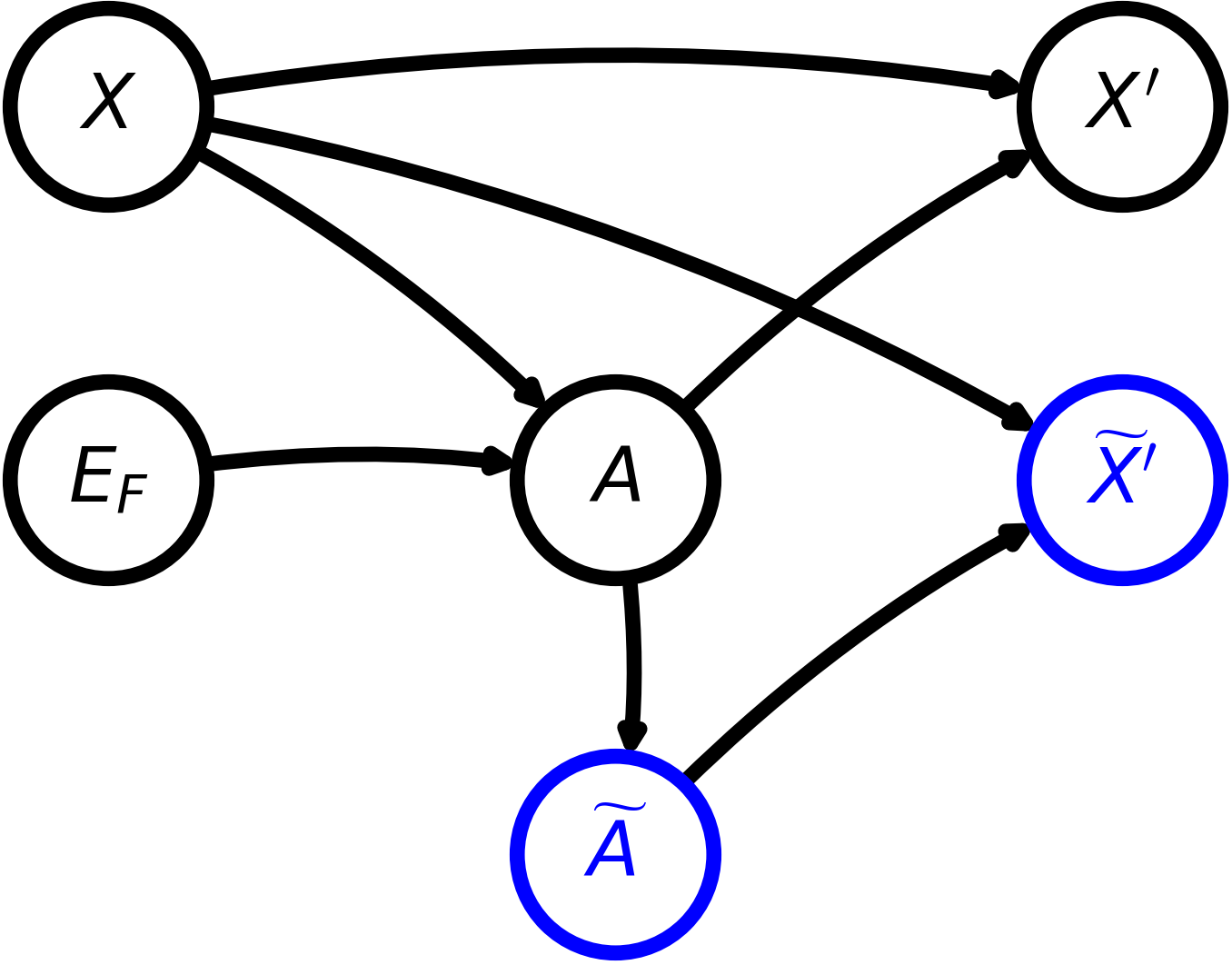}
  \caption{A structural causal model for the generation of modified data variables.}
  \label{fig:causal}
\end{figure}

\paragraph{Variables and the data-generation graph.}
We formalize the dependency structure using the structural causal model in Fig.~\ref{fig:causal}.
Here $X$ is the original input, $Y$ is the label, and $A$ is the attribution map.
To reason about ``information about the model/decision function'', we introduce an abstract random variable
\begin{equation}
  Z \;\triangleq\; (F,E),
\end{equation}
where $F$ denotes the (possibly random) decision function (e.g., trained parameters) and $E$ denotes the explainer identity (or, more generally, any model-side information accessible to the explainer).
For example, in the empirical setting below, $F$ can be the trained ResNet classifier and $E$ can indicate whether the explainer is Input-Gradient, Integrated Gradients, or Grad-CAM; conditioning on a fixed input $X=x$, $I(Z;A\mid X=x)$ captures the extent to which the produced map still varies with such model-side choices.
The attribution is generated as
\begin{equation}
  A \;=\; e_E(X,F) \quad (\text{possibly with internal randomness}),
\end{equation}
and ROAR constructs a binary mask $M_t$ by thresholding $A$ at drop rate $t$ (cf.\ Algorithm~\ref{algorithm:roar}),
then forms the modified input
\begin{equation}
  X_t^\prime \;=\; \phi(X,M_t) \;=\; (1-M_t)\odot X.
\end{equation}

\paragraph{Model/data-agnostic post-processing.}
We consider a post-processing map $k(\cdot)$ applied to the attribution:
\begin{equation}
  \widetilde{A} \;=\; k(A,U),
\end{equation}
where $U$ is an independent random seed (optional; deterministic $k$ is the special case without $U$).
Crucially, ``agnostic'' means that $k$ does not access $(X,Y,F,E)$ except through $A$.
ROAR is then run with $\widetilde{A}$ in place of $A$, producing $\widetilde{M}_t$ and $\widetilde{X}_t^\prime$.

\paragraph{Why conditioning on $X$ matters.}
In Fig.~\ref{fig:causal}, $A$ is influenced by both $X$ and $Z$, so marginal statements like $I(Z;A)$ can be confounded by the shared cause/collider structure.
To isolate the information carried by the attribution \emph{about} $Z$ for a fixed input instance, the right object is the conditional mutual information $I(Z;A\mid X)$.

\begin{theorem}[Conditional data processing for agnostic post-processing]
\label{theorem:dpi}
Assume $\widetilde{A}=k(A,U)$ where $U \perp (X,Y,Z,A)$ and $k$ uses $(A,U)$ only.
Then, for the causal graph in Fig.~\ref{fig:causal},
\begin{equation}
  I(Z;\widetilde{A}\mid X) \;\le\; I(Z;A\mid X).
\end{equation}
Moreover, for any (possibly randomized) measurable mapping $\psi$,
\begin{equation}
  I\!\bigl(Z;\psi(X,\widetilde{A})\mid X\bigr) \;\le\; I(Z;\widetilde{A}\mid X),
\end{equation}
and in particular,
\begin{gather}
  I(Z;\widetilde{M}_t\mid X) \;\le\; I(Z;\widetilde{A}\mid X),
  \tag{8a}\\
  I(Z;\widetilde{X}_t^\prime\mid X) \;\le\; I(Z;\widetilde{A}\mid X).
  \tag{8b}
\end{gather}
\end{theorem}

\noindent The proof is deferred to Appendix~\ref{app:proof_dpi}.

\paragraph{Interpretation.}
Theorem~\ref{theorem:dpi} states a one-way guarantee:
\emph{agnostic post-processing cannot increase the information the attribution carries about the model-side variable $Z$ for a fixed input}.
This is the correct DPI-based sanity check.
Importantly, DPI \emph{does not} imply any ordering such as
$I(Z;X_t^\prime\mid X)\ge I(Z;\widetilde{X}_t^\prime\mid X)$,
because $X_t^\prime=\phi(X,h_t(A))$ and $\widetilde{X}_t^\prime=\phi(X,h_t(\widetilde{A}))$ are generally \emph{two different} nonlinear transformations of $A$ (one does not have to be a processing of the other).
Therefore, a ROAR improvement after applying $k$ is not ruled out by DPI.

\begin{theorem}[ROAR can be improved while destroying model/explainer information]
\label{theorem:counterexample}
There exist random variables $(X,Y,Z)$, an attribution rule producing $A$, and an agnostic post-processing $k$ such that
\begin{gather}
  I(Z;\widetilde{A}\mid X)=0 \quad\text{but}\notag\\
  I(\widetilde{X}_t^\prime;Y) < I(X_t^\prime;Y)
\end{gather}
for a fixed drop rate $t$.
Consequently, an attribution pipeline can obtain a \emph{strictly better} ROAR score (i.e., smaller $I(\cdot;Y)$) even though the post-processed attribution contains \emph{no} information about $Z$.
\end{theorem}

\begin{proof}
Let $X=(X_1,X_2)$ with $X_1,X_2 \stackrel{\text{i.i.d.}}{\sim}\mathrm{Bernoulli}(1/2)$ and let $Y=X_1$.
Let $Z\in\{1,2\}$ be uniform and independent of $X$ (think of $Z$ as indexing ``what the explainer/model claims is important'').
Define the attribution map (two ``pixels'') by
\begin{equation}
  A \;=\;
  \begin{cases}
    (2,1) & \text{if } Z=1,\\
    (1,2) & \text{if } Z=2.
  \end{cases}
\end{equation}
Thus $A$ perfectly reveals $Z$, so $I(Z;A\mid X)=H(Z)=1$ bit.

Let $t=50\%$ and define ROAR's mask operator $h_t$ to select the coordinate whose attribution exceeds the empirical $50$-th percentile (no ties occur below).
Then $M_t=h_t(A)$ removes feature $Z$ and keeps the other one, hence
\begin{equation}
  X_t^\prime=
  \begin{cases}
    (0,X_2) & \text{if } Z=1,\\
    (X_1,0) & \text{if } Z=2.
  \end{cases}
\end{equation}
When $Z=2$, $X_t^\prime$ retains $X_1=Y$ and therefore $I(X_t^\prime;Y\mid Z=2)=H(Y)=1$ bit; in particular $I(X_t^\prime;Y)>0$.

Now define an agnostic post-processing $k$ (deterministic) by
\begin{equation}
  \widetilde{A} \;=\; k(A) \;\triangleq\; (A_1+A_2,\; A_1+A_2-\varepsilon),
\end{equation}
for any fixed $\varepsilon\in(0,1)$.
Since $A_1+A_2=3$ for both possible values of $A$, we have $\widetilde{A}\equiv(3,3-\varepsilon)$ deterministically, hence $I(Z;\widetilde{A}\mid X)=0$.

Because $\widetilde{A}_1>\widetilde{A}_2$, the $50\%$-drop mask $\widetilde{M}_t=h_t(\widetilde{A})$ always removes the first coordinate, so
\begin{equation}
  \widetilde{X}_t^\prime \;=\; (0,X_2).
\end{equation}
But $Y=X_1$ is independent of $X_2$, so $I(\widetilde{X}_t^\prime;Y)=0$.
Therefore $I(\widetilde{X}_t^\prime;Y) < I(X_t^\prime;Y)$ while $I(Z;\widetilde{A}\mid X)=0$, proving the claim.
\end{proof}

\paragraph{Consequence for ROAR.}
Theorem~\ref{theorem:counterexample} shows that, even under the strongest DPI sanity check (post-processing cannot add information about $Z$), the ROAR objective $I(X_t^\prime;Y)$ can strictly decrease after applying a $Z$-agnostic transformation $k$.
This means that a \emph{better} ROAR score does not necessarily certify that the attribution is \emph{more informative} about the model/decision function; it may instead reflect how the mask-generation pipeline interacts with the data distribution.

\paragraph{Our target phenomenon.}
Motivated by Theorem~\ref{theorem:dpi} and Theorem~\ref{theorem:counterexample}, we test for the following behavior in realistic vision settings: there exists a simple, model/data-agnostic post-processing $k$ that improves ROAR (reduces $I(X_t^\prime;Y)$, hence reduces re-trained accuracy) while, by DPI, never increasing the information about $Z$.

\begin{conjecture}
There exists a post-processing function $k$ such that, for some drop rate $t$,
\begin{gather}
  I(\widetilde{X}_t^\prime;Y) \;<\; I(X_t^\prime;Y),
  \tag{17a}\\
  \text{while}\quad
  I(Z;\widetilde{A}\mid X) \;\le\; I(Z;A\mid X).
  \tag{17b}
\end{gather}
\label{conjecture:ours}
\end{conjecture}

\newlength{\imgw}
\setlength{\imgw}{0.18\columnwidth}
\newcommand{\fimg}[1]{%
  \includegraphics[width=\imgw]{#1}%
}

\begin{figure}[t]
  \centering
  \setlength{\tabcolsep}{3pt}
  \renewcommand{\arraystretch}{1.05}

  \begin{adjustbox}{width=\columnwidth,center}
    \begin{tabular}{@{}
      >{\centering\arraybackslash}m{\imgw}
      !{\hspace{4pt}\vrule\hspace{4pt}}
      >{\centering\arraybackslash}m{0.08\columnwidth}
      !{\hspace{4pt}\vrule\hspace{4pt}}
      >{\centering\arraybackslash}m{\imgw}
      >{\centering\arraybackslash}m{\imgw}
      >{\centering\arraybackslash}m{\imgw}
      @{}}
      \toprule
      & & \small Plain & \small Max-pool & \small Gaussian \\
      \cmidrule(lr){3-5}

      \multirow{2}{*}{\includegraphics[width=\imgw]{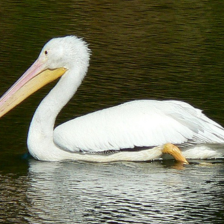}}
        & $\widetilde{m}$
        & \fimg{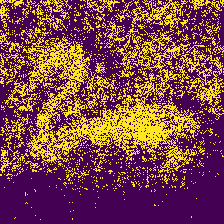}
        & \fimg{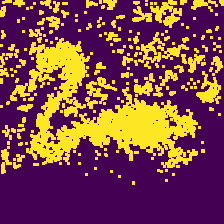}
        & \fimg{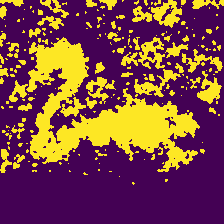} \\

      & $\widetilde{x}^{\prime}$
        & \fimg{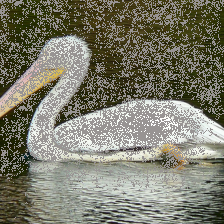}
        & \fimg{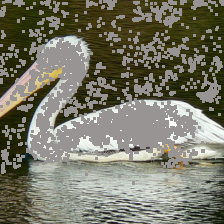}
        & \fimg{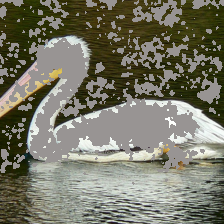} \\
      \bottomrule
    \end{tabular}
  \end{adjustbox}

  \caption{Model/data-agnostic attribution post-processings. The term `Plain' indicates no post-processing, i.e., $k(a)=a$. The leftmost image represents the original input image $x$. The feature importance measure used in this illustration is input-gradient.}
  \label{fig:gauss-max-pool}
\end{figure}

\subsection{Instantiation of Post-processing}
\label{sec:post-processing}

We now instantiate $k(\cdot)$ with two simple post-processings that are \emph{agnostic} in the sense of Theorem~\ref{theorem:dpi}:
(i) a Gaussian filter and (ii) a max-pooling (maximum) filter, both applied directly to the attribution map $a$.
These operations only transform the attribution values on the attribution grid and do not access the input $x$, the decision function $f(\cdot;\theta)$, or the explainer identity beyond the produced map itself.
Examples are shown in Fig.~\ref{fig:gauss-max-pool}.
Thus, the plain pipeline uses $A$, whereas the processed pipeline uses $\widetilde{A}=k(A)$, exactly matching the setup of Theorem~\ref{theorem:dpi}.
When the processed pipeline obtains lower ROAR/ROAD accuracy, it is the empirical analogue of Eq.~(17a), while Eq.~(17b) holds by construction of the agnostic $k$.
We do not estimate these mutual informations directly; re-trained accuracy is used as the observable ROAR/ROAD proxy.

The purpose of these choices is not to propose a better explainer, but to operationalize Conjecture~\ref{conjecture:ours}:
if such $k$ consistently improves ROAR/ROAD scores, then the benchmark can be optimized by procedures that, by DPI, cannot add model-side information.

\begin{figure}[t]
  \centering
  \begin{tabular}{c c c}
    \includegraphics[width=0.27\columnwidth]{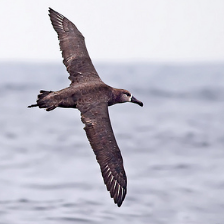} &
    \includegraphics[width=0.27\columnwidth]{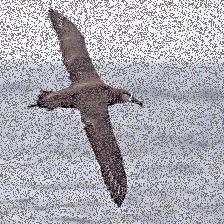} &
    \includegraphics[width=0.27\columnwidth]{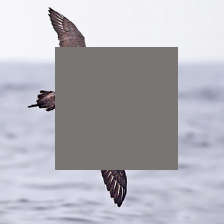}
  \end{tabular}
  \caption{(a) Input Image, (b) PixelRandom, and (c) BlockRandom. Examples of \textit{PixelRandom} and \textit{BlockRandom}.}
  \label{fig:random}
\end{figure}

\paragraph{Rationale of choice: from PixelRandom to BlockRandom.}
Our choice is motivated by a simple observation about mask geometry.
Consider two model-agnostic baselines:
\textit{PixelRandom} erases uniformly random pixels covering $t$ percent of the mask area, while \textit{BlockRandom} erases a single random rectangle covering $t$ percent of the area (Fig.~\ref{fig:random}).
Although both are uninformative about the model, BlockRandom can remove more structured content in natural images, leading to a larger reduction in class-relevant information.
This intuition is reflected by the ROAR results in Table~\ref{table:random-cub}.

\begin{table}[t]
  \centering
  \caption{The results of the ROAR method applied to the CUB-200 dataset using PixelRandom and BlockRandom. Each column represents the attribution drop rate $t \in T$.}
  \label{table:random-cub}
  \scalebox{0.8}{
  \begin{tabular}{c|ccccc}
    \toprule
                & 10\%     & 30\%     & 50\%     & 70\%     & 90\%     \\ \hline
    PixelRandom & 0.7556 & 0.7269 & 0.7099 & 0.6772 & 0.6195 \\
    BlockRandom & 0.6586 & 0.4384 & 0.2815 & 0.1705 & 0.0533 \\ \bottomrule
  \end{tabular}
  }
\end{table}

This motivates testing blur-inducing transformations (Gaussian smoothing and max-pooling) that tend to produce more spatially coherent (``block-like'') masks than the unprocessed attribution.
If ROAR rewards such mask shapes, then these agnostic transformations may improve ROAR/ROAD scores despite (by Theorem~\ref{theorem:dpi}) never increasing information about $Z$.

\begin{figure*}[t!]
\centering
\subfigure[CIFAR-10]{\includegraphics[width=0.9\textwidth]{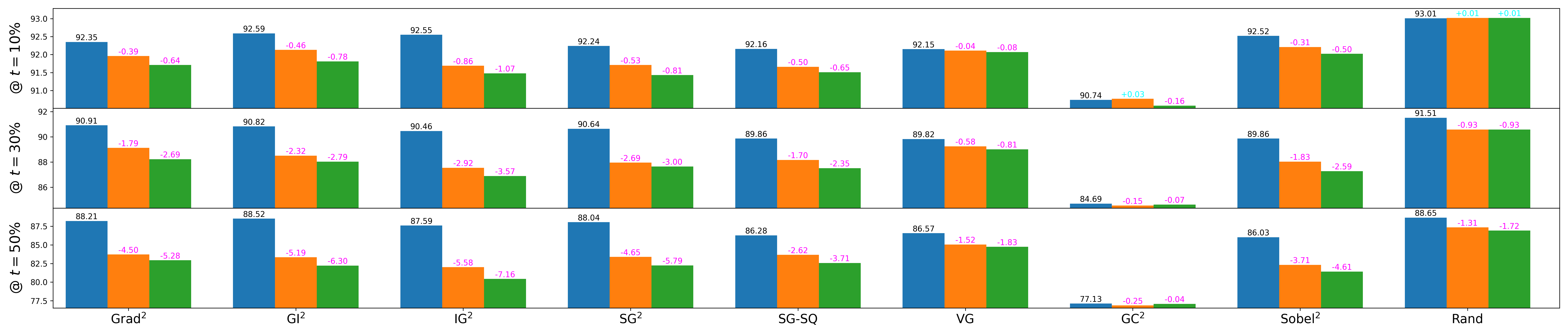}}\\
\subfigure[SVHN]{\includegraphics[width=0.9\textwidth]{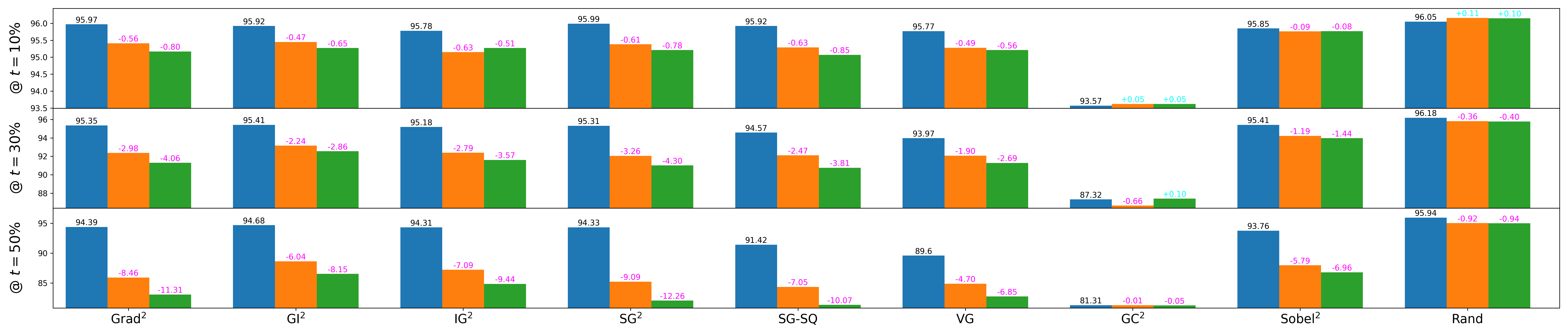}}\\
\subfigure[CUB-200]{\includegraphics[width=0.9\textwidth]{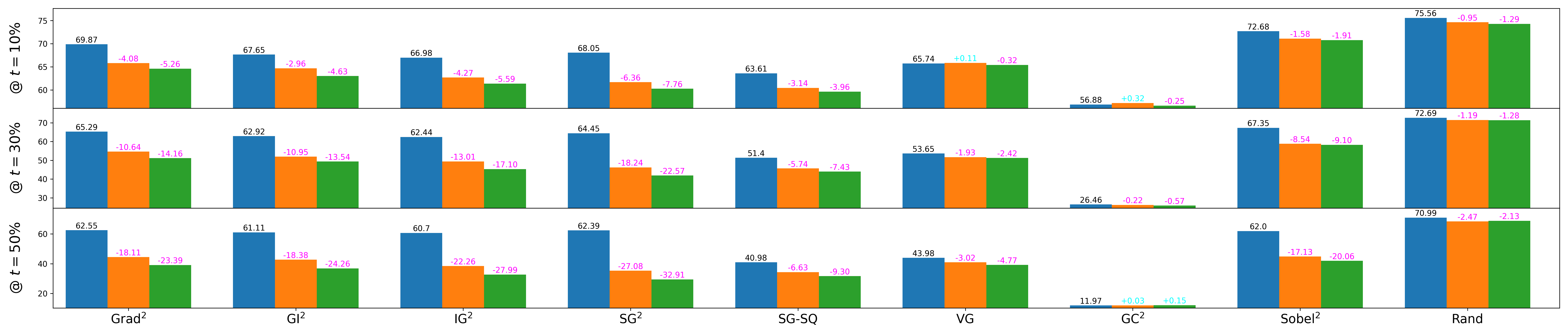}}\\
\subfigure{\includegraphics[width=0.3\textwidth]{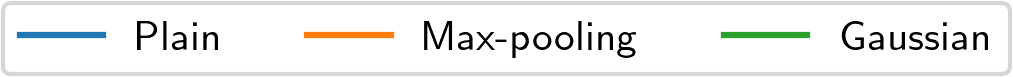}}
\caption{The effect of Gaussian filtering and max-pooling on the ROAR metric. The labels `P', `M', and `G' refer to `Plain', `Max-pooling', and `Gaussian filter', respectively. The numbers at the top of each column indicate the attribution drop rate. For ease of interpretation, the results of max-pooling and Gaussian filtering are expressed as the difference from the plain method. A decrease in model accuracy is indicated by a magenta `$-$' sign, while an increase is represented by a cyan `$+$' sign. For optimal viewing, it is recommended to zoom in.}
\label{fig:effects-of-processing}
\end{figure*}

\section{Experiments}
\label{sec:exps}

\subsection{Experimental Settings}
We perform a series of experiments on three image classification datasets as \cite{hooker2019benchmark, rong2022consistent} do: CIFAR-10~\citep{krizhevsky2009learning}, SVHN~\citep{netzer2011reading}, and CUB-200~\citep{WelinderEtal2010}. For our neural network implementation, we utilize the PyTorch model zoo implementation of ResNet18~\citep{he2016deep} with a customized kernel size for CIFAR-10 and SVHN, and ResNet50 for CUB-200.

\begin{figure*}[t!]
\centering
\subfigure[CIFAR-10]{\includegraphics[width=0.9\textwidth]{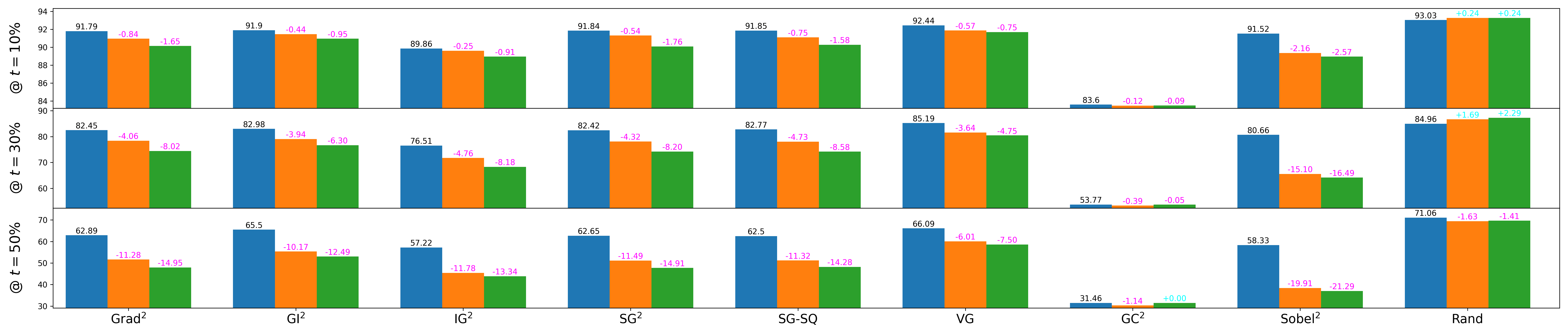}}\\
\subfigure[SVHN]{\includegraphics[width=0.9\textwidth]{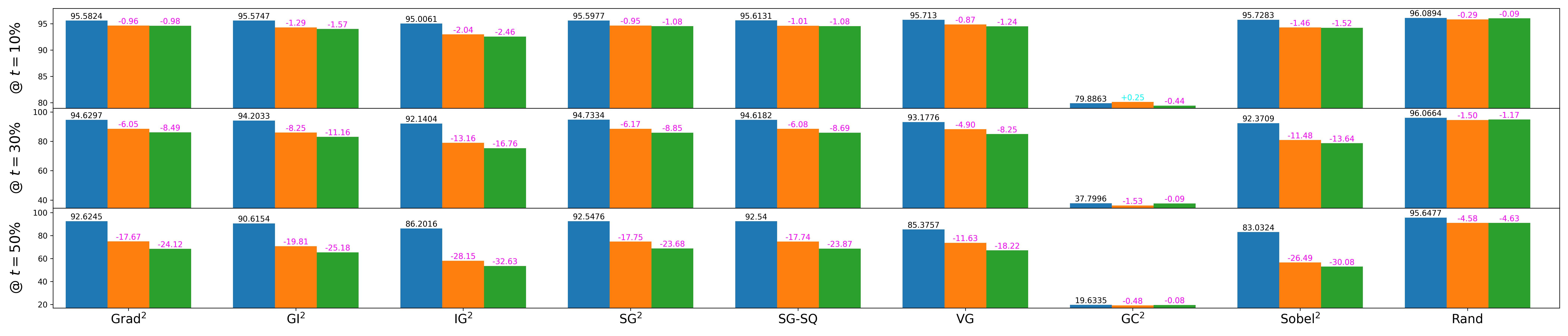}}\\
\subfigure[CUB-200]{\includegraphics[width=0.9\textwidth]{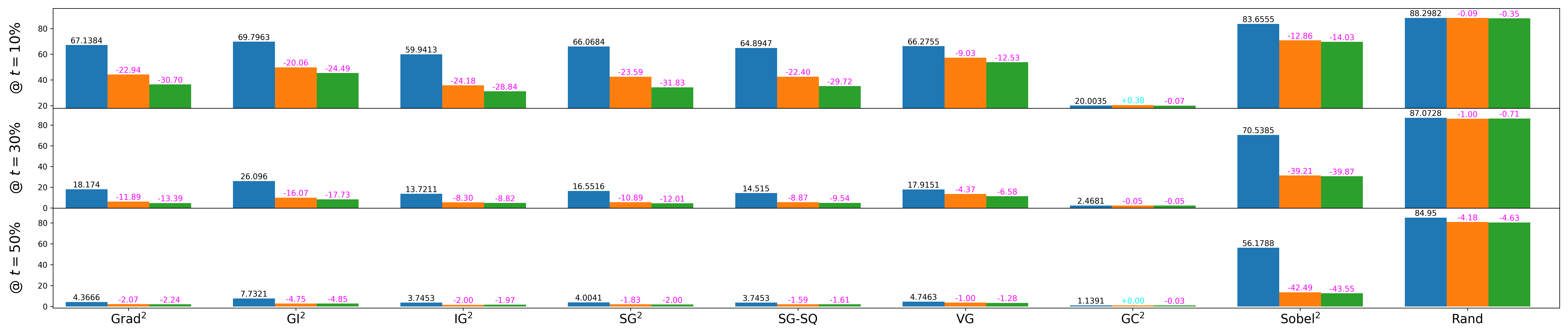}}\\
\subfigure{\includegraphics[width=0.3\textwidth]{figure/0/legend.png}}
\caption{The effect of Gaussian filtering and max-pooling on the ROAD metric. The conventions used in this figure are the same as those in Figure \ref{fig:effects-of-processing}, except that this figure relates to the ROAD benchmark instead of the ROAR benchmark.}
\label{fig:effects-of-processing-road}
\end{figure*}

We compare the ROAR performance of several feature importance estimates with a fixed neural network architecture on each dataset. These estimates include Input-Gradient~\citep{JMLR:v11:baehrens10a,simonyan2014deep} (Grad), Grad * Input~\citep{shrikumar2017learning} (GI), Integrated Gradient~\citep{sundararajan2017axiomatic} (IG), SmoothGrad~\citep{smilkov2017smoothgrad} (SG), VarGrad~\citep{adebayo2018local} (VG), and Grad-CAM~\citep{selvaraju2017grad} (GC). Following~\cite{hooker2019benchmark}, we primarily report a squared version of these attribution methods, abbreviated as Grad$^2$, GI$^2$, IG$^2$, SG$^2$, VG, and GC$^2$. It should be noted that squaring does not apply to VG, as it is itself a squaring method. Additionally, we include SG-SQ, the pre-squared version of SmoothGrad~\citep{hooker2019benchmark}, as a comparative methodology.

For reference, we also report the plain, non-squared version of Input-Gradient (Grad), PixelRandom (Rand), and the squared version of Sobel edge detection (Sobel$^2$) as baselines. To simplify our results, we use attribution drop rates of 10\%, 30\%, and 50\% in the ROAR protocol. To account for trial variability, all results are reported as the average of five trials. Further experimental details can be found in Appendix~\ref{app:exp_details}.

\subsection{Effects of Agnostic Post-processing}
\label{sec:exp-postprocessing}

To examine the effect of agnostic post-processing on the ROAR metric, we analyze its impact across datasets and attribution methods. The results in Figure~\ref{fig:effects-of-processing} show that max-pooling and Gaussian filtering frequently lower the re-trained accuracy, which corresponds to a better ROAR score.

The benefits of post-processing are not uniform and depend on the dataset, attribution method, and drop rate. In our experiments, post-processing was particularly effective on the CUB-200 dataset compared to the other two datasets.
Max-pooling and Gaussian filtering generally yielded favorable ROAR scores, while the GC$^2$ method produced inconclusive results for all datasets and drop rates. The unusual behavior of GC$^2$ is covered in Section~\ref{sec:exp-rel}. These results support Conjecture~\ref{conjecture:ours}: agnostic transformations can obtain better ROAR scores without adding information about the model-side variable $Z$.
Because individual bar differences can be small, Table~\ref{tab:directional-summary} summarizes the directional pattern over the plotted mean accuracies.

\begin{table}[t]
  \centering
  \caption{Number of mean comparisons in which post-processing lowers the final accuracy relative to the plain attribution map. Lower accuracy is better under ROAR/ROAD. Each cell aggregates $3$ datasets $\times$ $9$ attribution/baseline maps $\times$ $3$ drop rates, for $81$ comparisons.}
  \label{tab:directional-summary}
  \scalebox{0.86}{
  \begin{tabular}{lcc}
    \toprule
    Benchmark & Max-pooling & Gaussian \\
    \midrule
    ROAR & $74/81$ & $76/81$ \\
    ROAD & $76/81$ & $78/81$ \\
    \bottomrule
  \end{tabular}}
\end{table}

\subsection{ROAD Analysis}
\label{sec:road}

We conducted a similar experiment on the ROAD benchmark (Figure~\ref{fig:effects-of-processing-road}), a variation of ROAR designed to reduce confounds in removal-based evaluation~\citep{rong2022consistent}. ROAD changes the debiasing/removal procedure, but it still constructs its score from attribution-derived masks and the resulting modified inputs. Therefore, the agnostic post-processing intervention considered here occurs upstream of the ROAD scoring step, and the DPI argument for $A \mapsto \widetilde{A}=k(A)$ applies in the same way.

As summarized in Table~\ref{tab:directional-summary}, max-pooling and Gaussian filtering lower the mean ROAD accuracy in most comparisons. This suggests that ROAD's debiasing step does not by itself eliminate the mask-geometry sensitivity studied in this paper, although the exact magnitude remains method- and dataset-dependent.

\subsection{Delving into Relationship between ROAR Benchmark and Blurriness}
\label{sec:exp-rel}
\begin{figure}[t!]
    \centering
\begin{tabular}{c c c}
\subfigure[Grad$^2$]{\includegraphics[width=0.28\columnwidth]{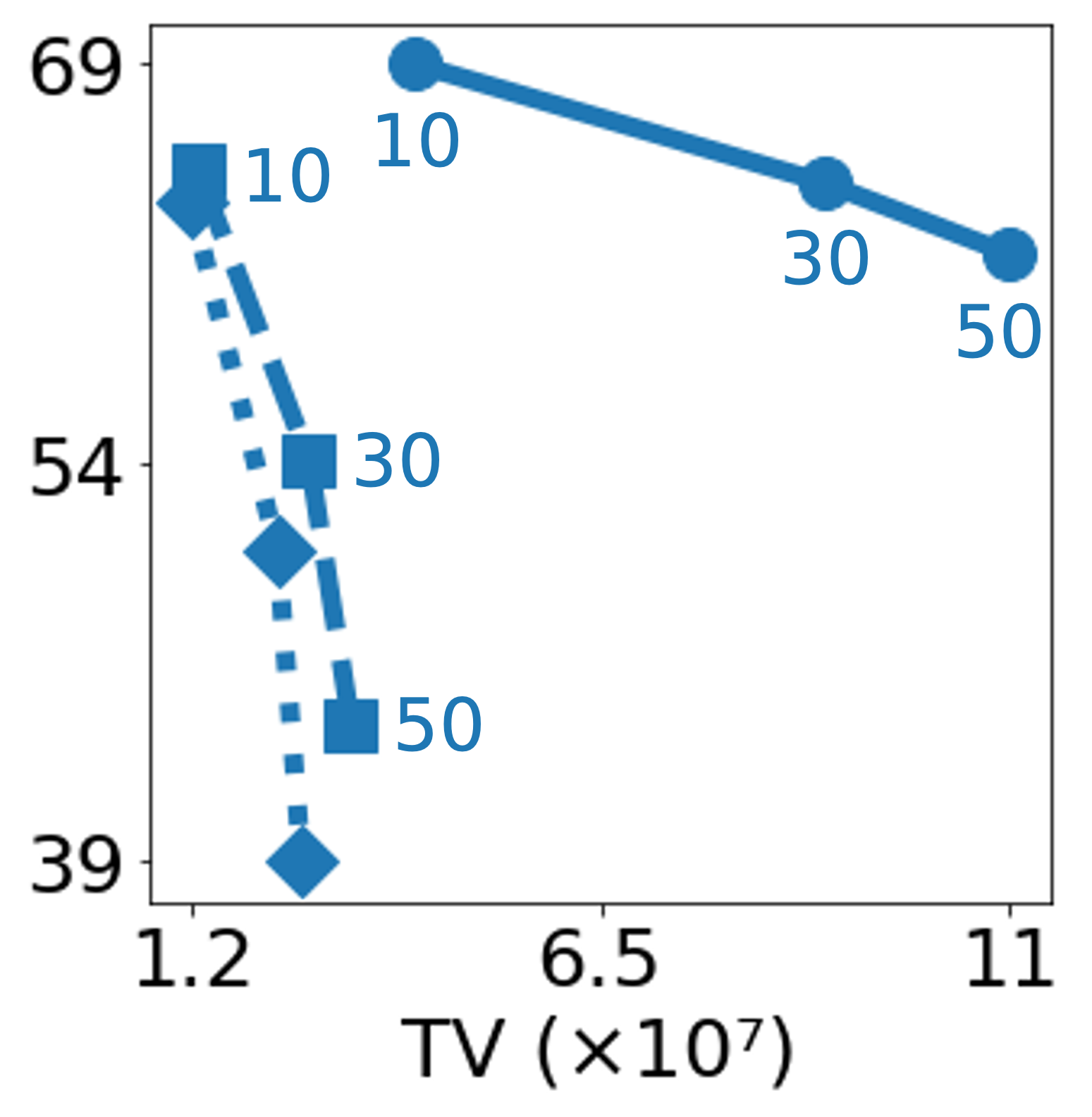}} &
\subfigure[GI$^2$]{\includegraphics[width=0.28\columnwidth]{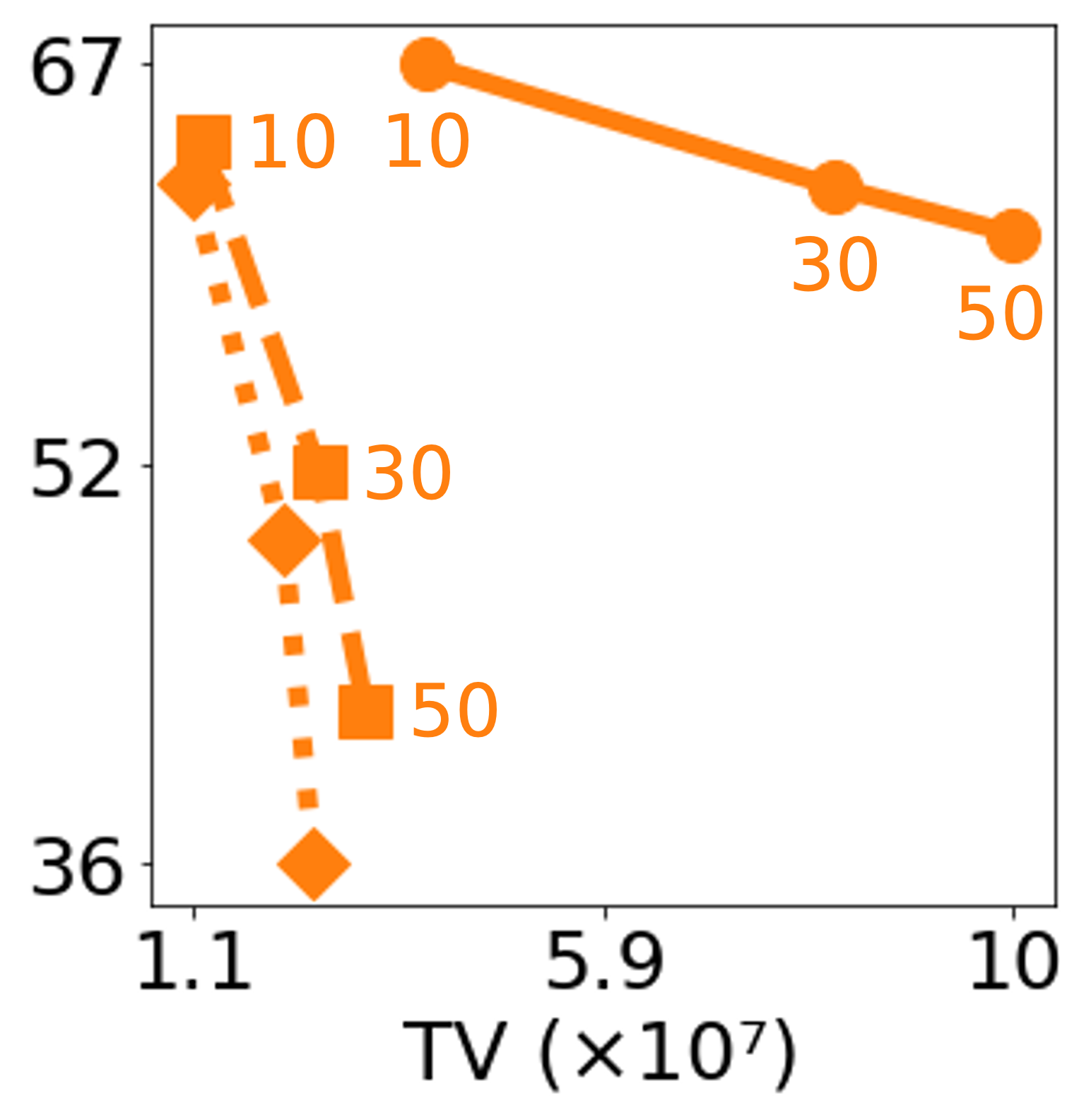}} &
\subfigure[IG$^2$]{\includegraphics[width=0.28\columnwidth]{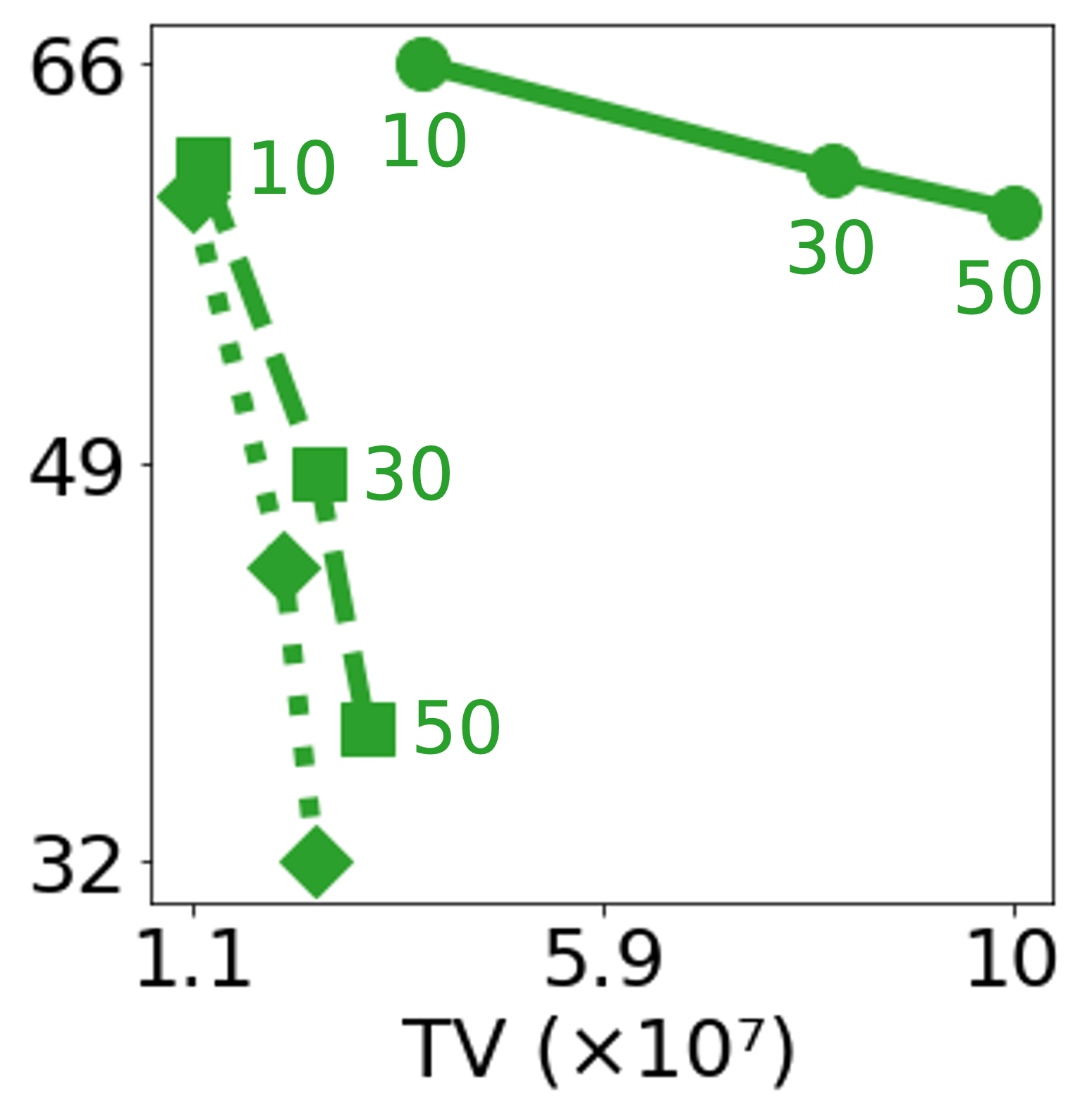}}\\
\subfigure[SG$^2$]{\includegraphics[width=0.28\columnwidth]{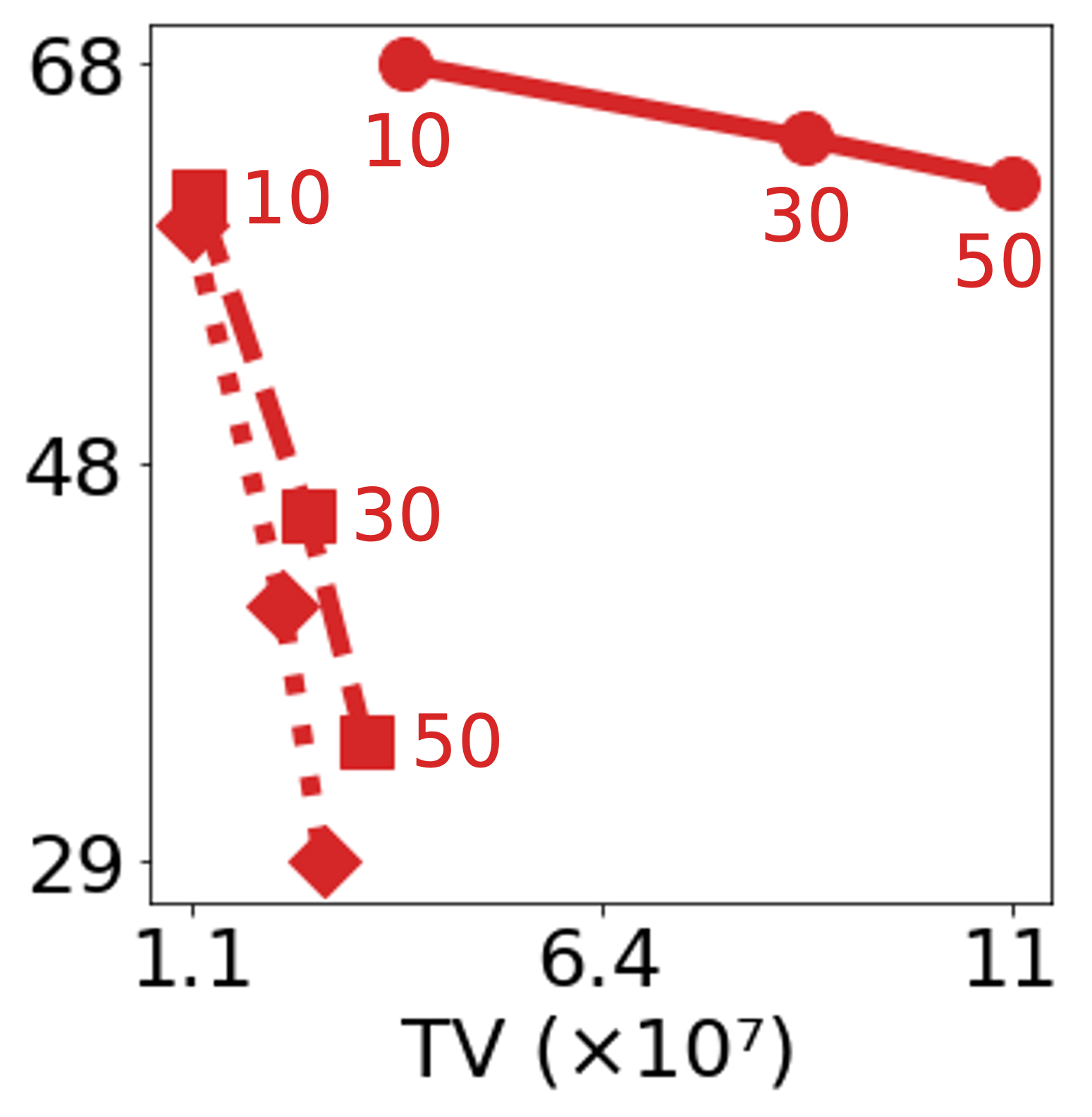}} &
\subfigure[SG-SQ]{\includegraphics[width=0.28\columnwidth]{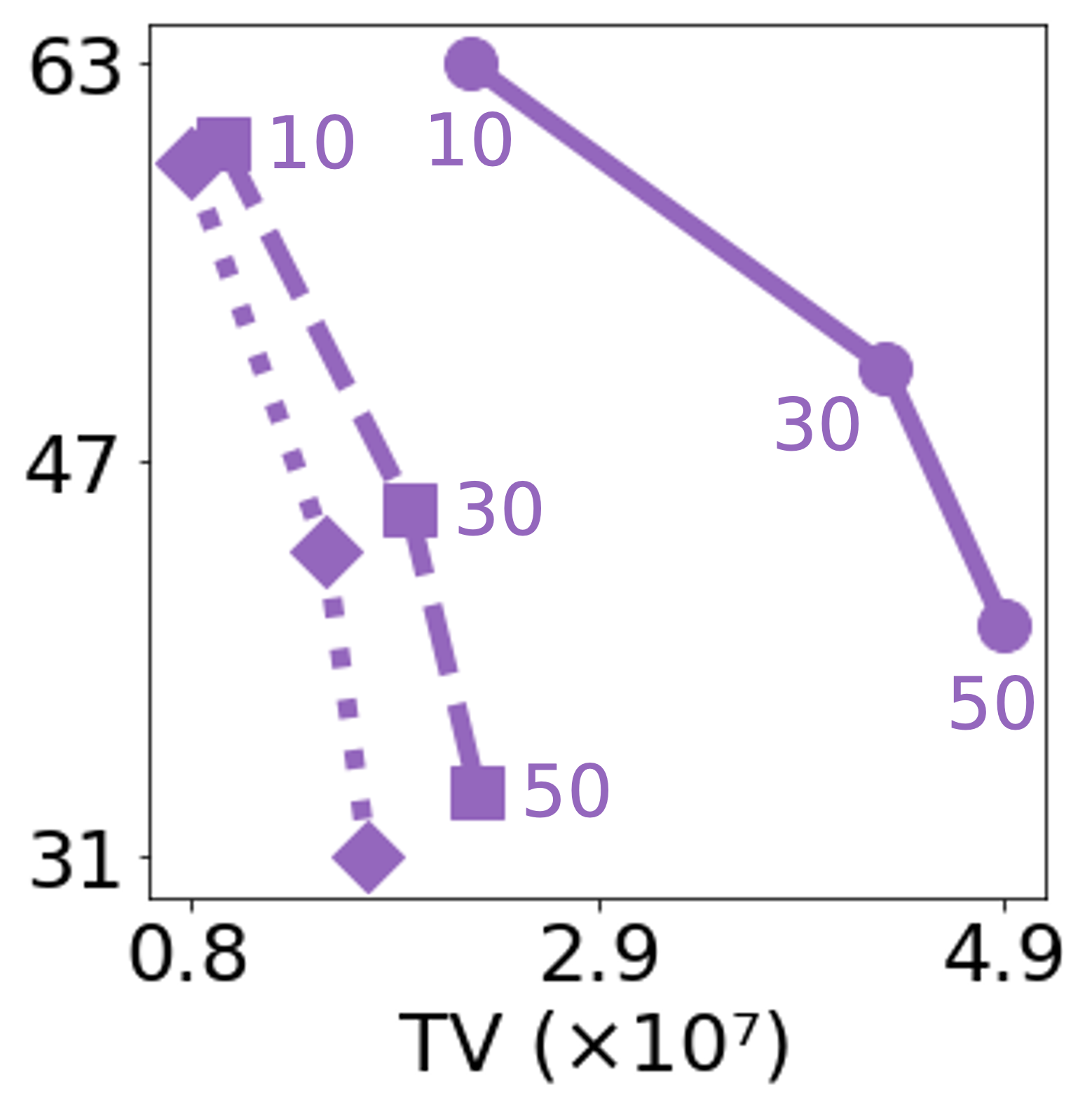}} &
\subfigure[VG]{\includegraphics[width=0.28\columnwidth]{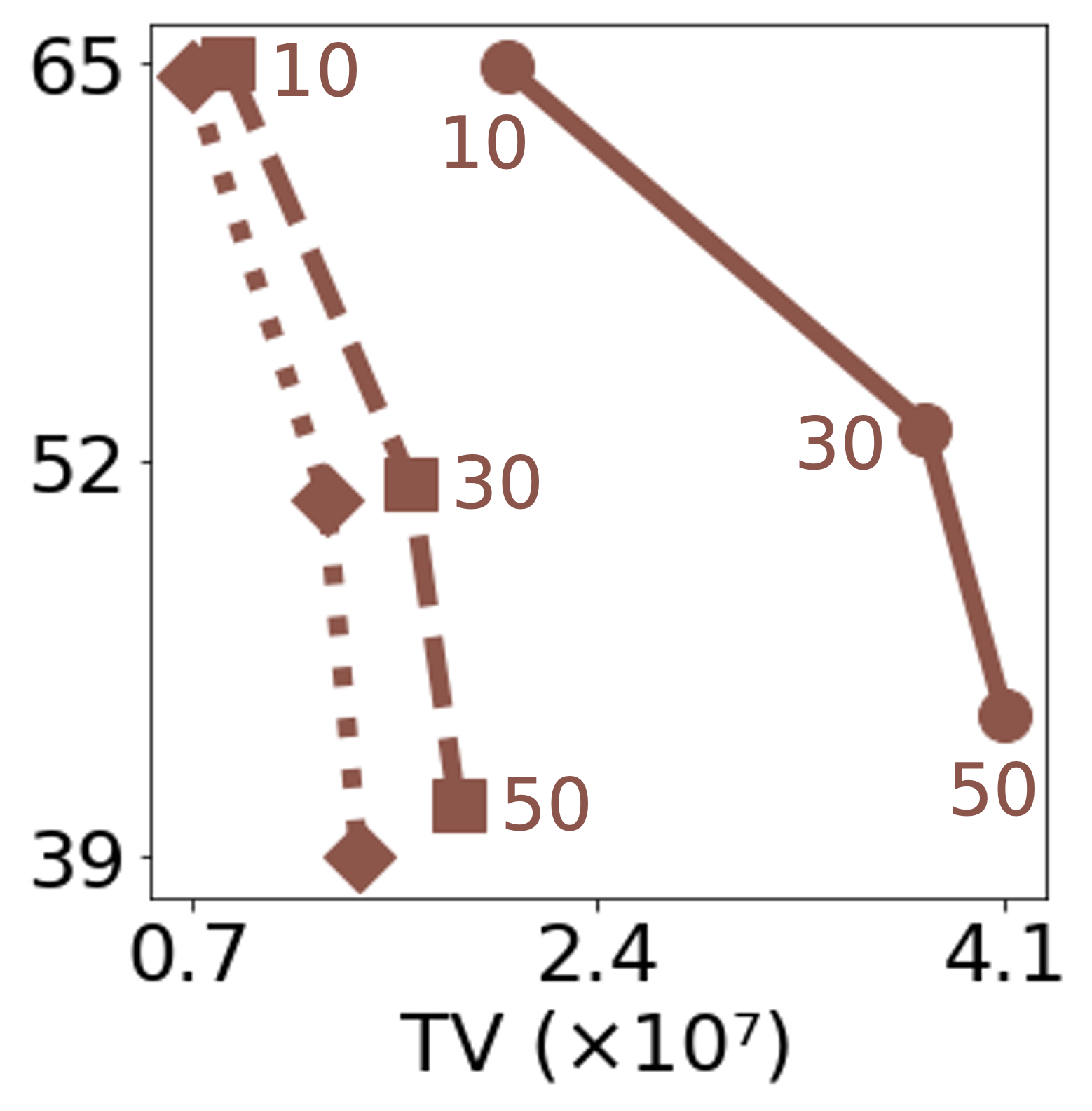}}\\
\subfigure[GC$^2$]{\includegraphics[width=0.28\columnwidth]{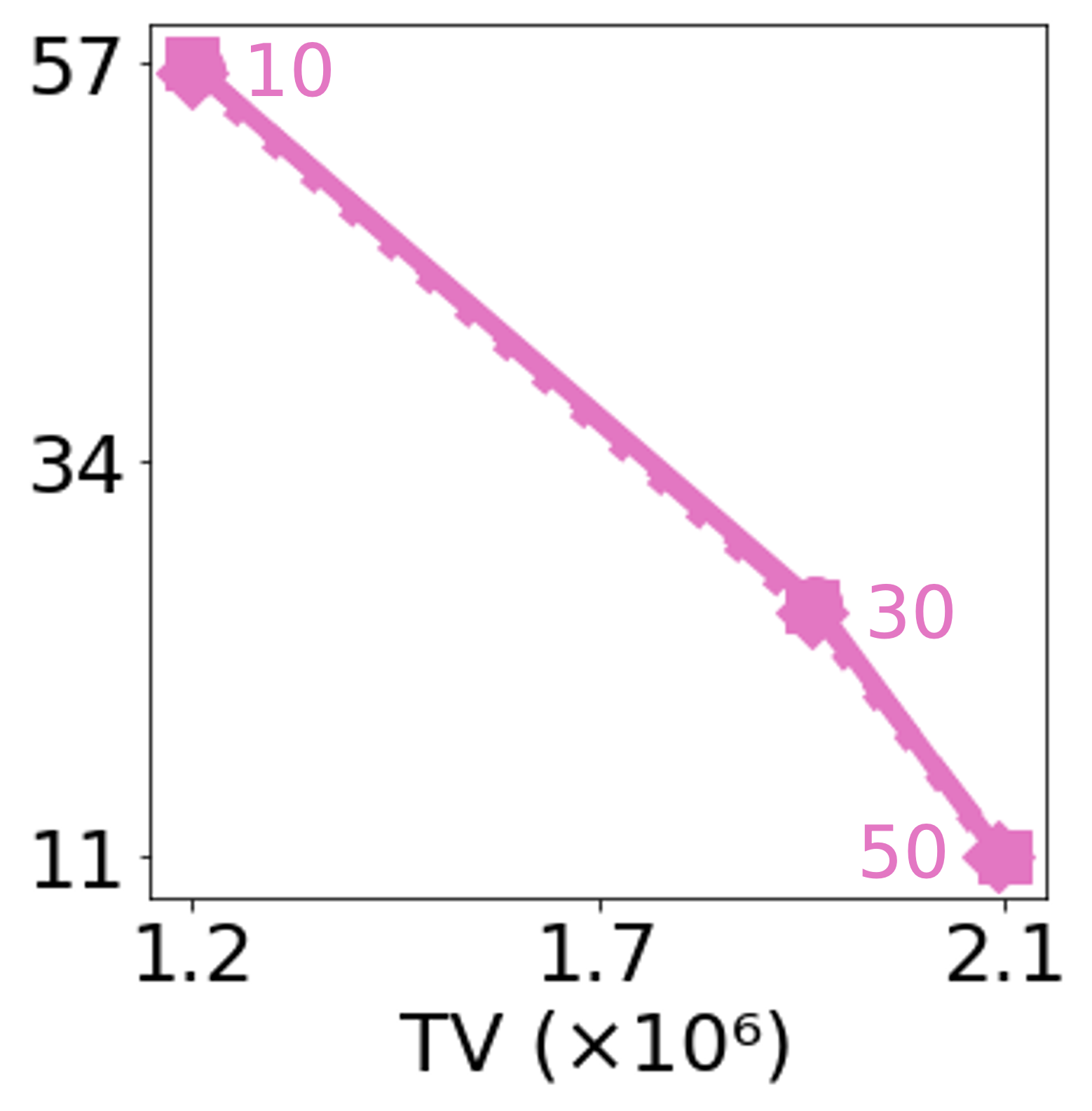}} &
\subfigure[Sobel$^2$]{\includegraphics[width=0.28\columnwidth]{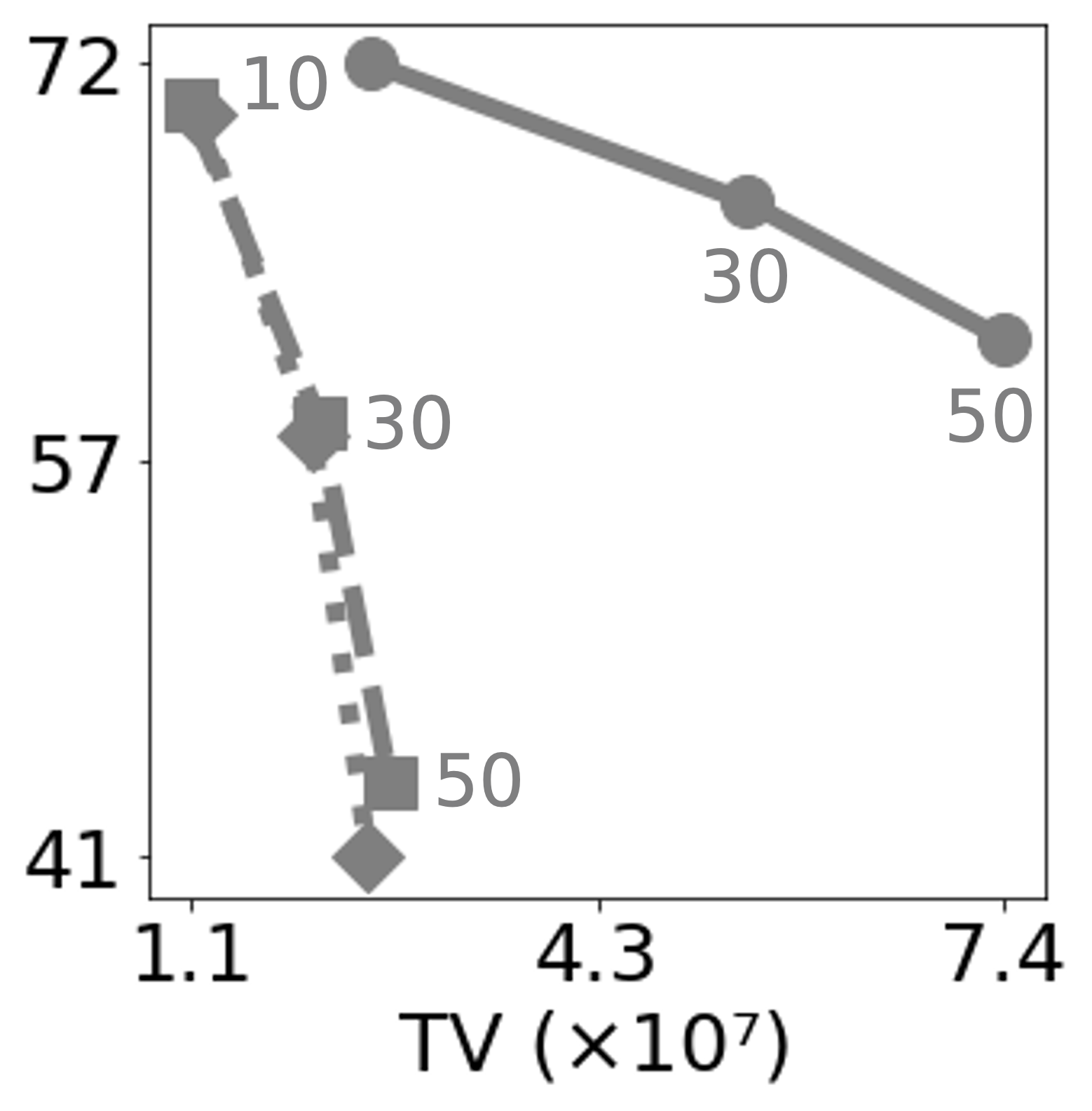}} &
\subfigure[Rand]{\includegraphics[width=0.28\columnwidth]{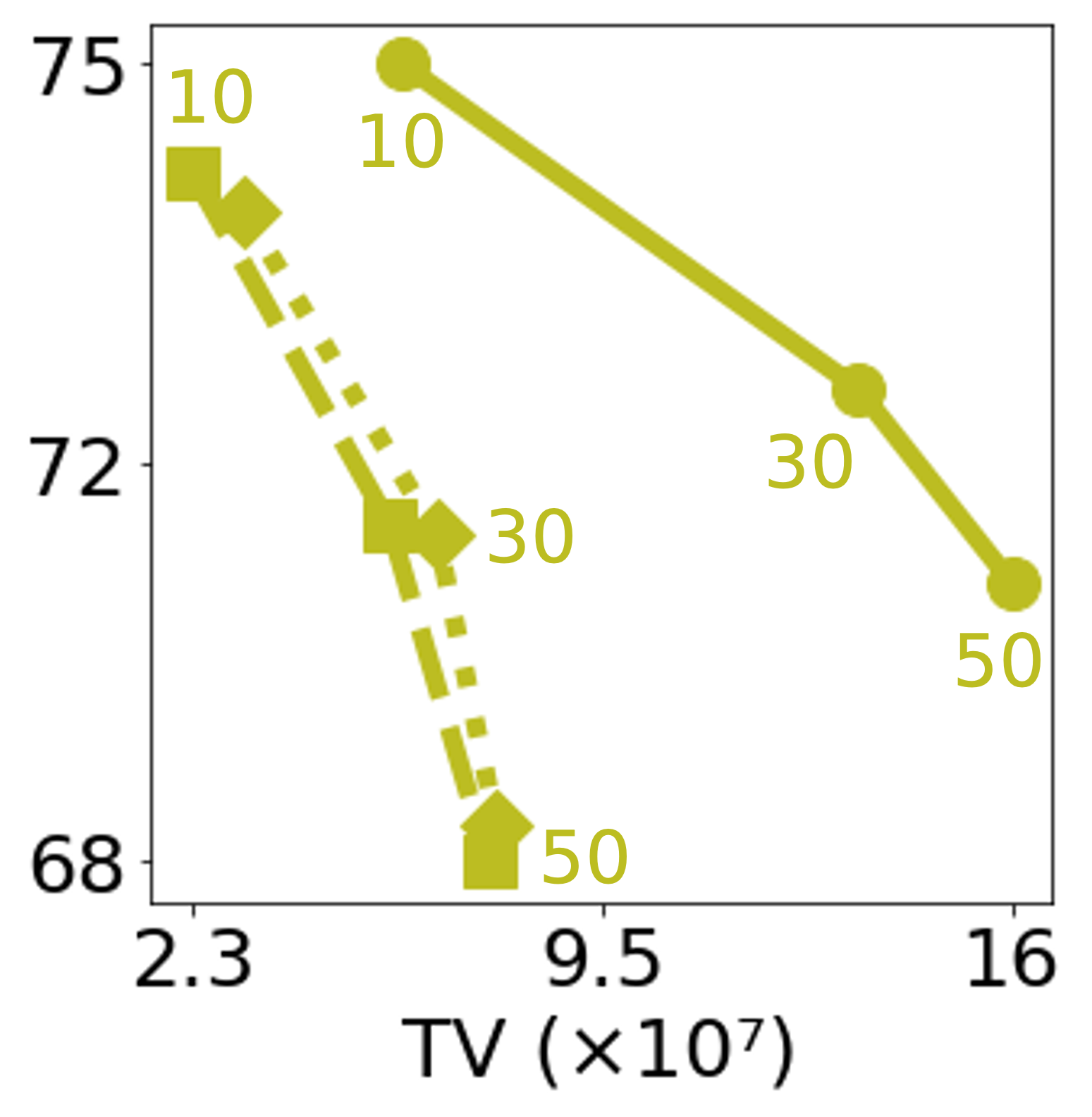}}
\end{tabular}\\
\subfigure{\includegraphics[width=0.7\columnwidth]{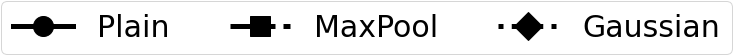}}
    \caption{Effects of Gaussian filtering and max-pooling on the total variation of attribution masks (TV) and model accuracy on the CUB-200 dataset. The number displayed above each point indicates the attribution drop rate. The y-axis represents the final test accuracy (\%).}
    \label{fig:tv-accuracy-cub200}
\end{figure}

Going further, we examine the relationship between the ROAR metric and attribution blurriness, as discussed in Section~\ref{sec:post-processing}. We measure blurriness through the average total variation (TV) of attribution masks for the training set $A_{train}$ in each ROAR trial; lower TV corresponds to blurrier masks. The results in Figure~\ref{fig:tv-accuracy-cub200} show that lower-TV attributions tend to yield lower final accuracy in the ROAR protocol. Results on CIFAR-10 and SVHN are reported in Figures~\ref{fig:tv-accuracy-cifar10} and~\ref{fig:tv-accuracy-svhn} in Appendix~\ref{app:plot}.

Our findings indicate that for most attribution methods and drop rates, max-pooling and Gaussian filtering reduce both the total variation and the model's performance. However, this trend does not hold for the GC$^2$ method, as its unique approach (derived from gradients on intermediate feature maps rather than back-propagation on inputs) results in attribution maps with a smaller resolution, which are less affected by the post-processing $k(\cdot)$. Similar patterns were observed in the CIFAR-10 and SVHN datasets (Figures~\ref{fig:tv-accuracy-cifar10} and~\ref{fig:tv-accuracy-svhn}).

Next, we examine the relationship between model performance and total variation without post-processing. Figure~\ref{fig:relation-sq} presents a scatter plot of all attribution methods' results for each dataset and drop rate. It shows a strong linear association between total variation and model accuracy, excluding results after post-processing. Table~\ref{tab:tv-r2} summarizes the corresponding $R^2$ values, and the non-squared attribution results in Figure~\ref{fig:relation} show a similar pattern. Thus, because lower accuracy is treated as better in ROAR, blurrier/lower-TV masks can be favored even when the blurring is introduced by an agnostic post-processing step.

\begin{table}[t]
  \centering
  \caption{Coefficient of determination ($R^2$) for the linear relationship between total variation of squared attribution masks and final ROAR test accuracy in Figure~\ref{fig:relation-sq}.}
  \label{tab:tv-r2}
  \scalebox{0.82}{
  \begin{tabular}{lccc}
    \toprule
    Dataset & $t=10\%$ & $t=30\%$ & $t=50\%$ \\
    \midrule
    CIFAR-10 & $0.84$ & $0.85$ & $0.82$ \\
    SVHN & $0.92$ & $0.90$ & $0.95$ \\
    CUB-200 & $0.91$ & $0.90$ & $0.92$ \\
    \bottomrule
  \end{tabular}}
\end{table}

\section{Related Works}
\label{related_works}

Evaluating the quality of attribution methods is an active area~\citep{nauta2022anecdotal}. \cite{doshi2017towards} categorize evaluation into functionally-grounded, human-grounded, and application-grounded approaches; \cite{zhou2021evaluating} further divide functionally-grounded evaluations into clarity, broadness, simplicity, completeness, and soundness. Our focus is on the ROAR protocol~\citep{hooker2019benchmark}, a widely-used soundness-oriented methodology.

ROAR and its variants have been used across image tasks~\citep{chefer2021transformer, yang2020learning}, NLP~\citep{ismail2021improving, zhang-etal-2021-sample}, graph analysis~\citep{funke2022z}, and time-series~\citep{10.1145/3394486.3403071, ismail2020benchmarking}, and ROAR has been adopted as the primary metric in \cite{meng2022interpretability}'s standard benchmark. Our research focuses on the intriguing and unintended behavior of the ROAR protocol.

\paragraph{Existing Discussions on the Limitations of ROAR.}
\cite{hooker2019benchmark} noted that feature redundancy can prevent ROAR from distinguishing meaningless attributions from redundant features, though no such case was observed in practice.
\cite{rong2022consistent} identified information leakage through mask shape and proposed RemOve-And-Debias (ROAD) to reduce confounding factors; coordinate/geometry sensitivity was further studied by~\citet{park2024geometric}. Our findings address a different issue, namely the dependency of the data generation process, rather than mask-based information leakage. As discussed in Section~\ref{sec:road}, ROAD still relies on attribution-derived masks and modified inputs, and our experiments suggest that its debiasing step does not eliminate sensitivity to agnostic post-processing. More recently, \citet{bora2025fries} proposed frameworks to estimate the inconsistency of saliency metrics themselves.

\section{Discussion}
\paragraph{Practical Takeaways.}
Our work does not aim to discourage the use of ROAR (or ROAD) in future research, but rather to emphasize the importance of exercising caution when employing and interpreting these methods. We present two main practical takeaways.

\begin{itemize}
\item Ranking attribution methods solely by ROAR scores can be misleading~\citep{canha2025functionally}. We recommend jointly reporting factors that contribute to ROAR bias, such as attribution total variation (Fig.~\ref{fig:relation-sq}).
\item ROAR trends are partly influenced by data structure, independent of model and explainer quality. Similar relationships between data structure and feature redundancy have been reported in time series~\citep{radovic2017minimum} and graph structures~\citep{liu2022rethinking}.
\end{itemize}

\paragraph{Implications for Mechanistic Interpretability.}
Attribution methods increasingly serve as components in broader mechanistic interpretability pipelines, including circuit analysis~\citep{wang2023interpretability}, sparse dictionary validation~\citep{bricken2023monosemanticity}, and safety-oriented monitoring of deployed models.
An evaluation metric with such confounds, if used as the sole criterion for selecting attributions, may propagate biases into downstream mechanistic conclusions.
As \citet{sharkey2025open} note, validating interpretability methods using benchmarks remains an important open problem. Our findings contribute a concrete diagnostic result clarifying what ROAR-style benchmarks can and cannot measure, and highlight the need for metrics that assess information about the \emph{decision function} more directly rather than relying only on proxy accuracy drops that may be confounded by mask geometry.

\paragraph{Extension and Limitations.}
The core issue, namely confounds in the data generation process, extends to other perturbation-based evaluation methods~\citep{samek2016evaluating, yeh2019fidelity, bhatt2020evaluating, ismail2020benchmarking}.
Our findings are primarily validated on image tasks where spatial structure is strongly expressed. Biases may manifest differently in text or graph domains, and exploring these settings remains an important direction for scaling interpretability evaluation.
Our theoretical analysis assumes the idealized causal model in Fig.~\ref{fig:causal}. In practice, retraining dynamics may introduce additional confounds beyond what DPI captures.

\section{Conclusion}
This paper highlights a limitation of using the ROAR metric and its variant as stand-alone evaluations of feature importance estimates. We show theoretically that attributions carrying \emph{less} information about the model can achieve better ROAR scores, and empirically observe that simple agnostic post-processing often improves ROAR/ROAD scores through a blurriness-related bias.
As the mechanistic interpretability community increasingly relies on attribution methods to validate hypotheses about model internals and to support safety-relevant applications, careful benchmark interpretation becomes important. We hope our results encourage the development of evaluation protocols.

\bibliographystyle{icml2026}
\bibliography{references}

@inproceedings{hooker2019benchmark,
  title={A benchmark for interpretability methods in deep neural networks},
  author={Hooker, Sara and Erhan, Dumitru and Kindermans, Pieter-Jan and Kim, Been},
  booktitle={NeurIPS},
  pages={9737--9748},
  year={2019}
}

@inproceedings{selvaraju2017grad,
  title={Grad-cam: Visual explanations from deep networks via gradient-based localization},
  author={Selvaraju, Ramprasaath R and Cogswell, Michael and Das, Abhishek and Vedantam, Ramakrishna and Parikh, Devi and Batra, Dhruv},
  booktitle={CVPR},
  pages={618--626},
  year={2017}
}

@inproceedings{adebayo2020debugging,
  author={Julius Adebayo and Michael Muelly and Ilaria Liccardi and Been Kim},
  title={Debugging Tests for Model Explanations},
  year={2020},
  cdate={1577836800000},
  url={https://proceedings.neurips.cc/paper/2020/hash/075b051ec3d22dac7b33f788da631fd4-Abstract.html},
  booktitle={NeurIPS}
}

@article{adebayo2018sanity,
  title={Sanity checks for saliency maps},
  author={Adebayo, Julius and Gilmer, Justin and Muelly, Michael and Goodfellow, Ian and Hardt, Moritz and Kim, Been},
  journal={NeurIPS},
  year={2018}
}

@article{smilkov2017smoothgrad,
  title={Smoothgrad: removing noise by adding noise},
  author={Smilkov, Daniel and Thorat, Nikhil and Kim, Been and Vi{\'e}gas, Fernanda and Wattenberg, Martin},
  journal={ICML Workshop},
  year={2017}
}

@inproceedings{shrikumar2017learning,
  title={Learning important features through propagating activation differences},
  author={Shrikumar, Avanti and Greenside, Peyton and Kundaje, Anshul},
  booktitle={ICML},
  year={2017}
}

@inproceedings{sundararajan2017axiomatic,
author = {Sundararajan, Mukund and Taly, Ankur and Yan, Qiqi},
title = {Axiomatic Attribution for Deep Networks},
year = {2017},
booktitle = {ICML},
pages = {3319–3328},
numpages = {10},
}

@inproceedings{kim2019saliency,
  title={Why are saliency maps noisy? cause of and solution to noisy saliency maps},
  author={Kim, Beomsu and Seo, Junghoon and Jeon, Seunghyeon and Koo, Jamyoung and Choe, Jeongyeol and Jeon, Taegyun},
  booktitle={ICCV Workshop},
  pages={4149--4157},
  year={2019},
  organization={IEEE}
}

@article{kim2019bridging,
  title={Bridging adversarial robustness and gradient interpretability},
  author={Kim, Beomsu and Seo, Junghoon and Jeon, Taegyun},
  journal={ICLR Workshop},
  year={2019}
}

@article{schramowski2020making,
  title={Making deep neural networks right for the right scientific reasons by interacting with their explanations},
  author={Schramowski, Patrick and Stammer, Wolfgang and Teso, Stefano and Brugger, Anna and Herbert, Franziska and Shao, Xiaoting and Luigs, Hans-Georg and Mahlein, Anne-Katrin and Kersting, Kristian},
  journal={Nature Machine Intelligence},
  year={2020},
}

@inproceedings{bhatt2020evaluating,
  title={Evaluating and aggregating feature-based model explanations},
  author={Bhatt, Umang and Weller, Adrian and Moura, Jos{\'e} MF},
  booktitle={IJCAI},
  year={2020}
}

@inproceedings{10.1145/3394486.3403071,
author = {Liang, Jian and Bai, Bing and Cao, Yuren and Bai, Kun and Wang, Fei},
title = {Adversarial Infidelity Learning for Model Interpretation},
year = {2020},
isbn = {9781450379984},
url = {https://doi.org/10.1145/3394486.3403071},
booktitle = {KDD},
pages = {286–296},
numpages = {11}
}

@article{o2020generative,
  title={Generative causal explanations of black-box classifiers},
  author={O'Shaughnessy, Matthew and Canal, Gregory and Connor, Marissa and Davenport, Mark and Rozell, Christopher},
  journal={NeurIPS},
  year={2020}
}

@inproceedings{yang2020learning,
  title={Learning propagation rules for attribution map generation},
  author={Yang, Yiding and Qiu, Jiayan and Song, Mingli and Tao, Dacheng and Wang, Xinchao},
  booktitle={ECCV},
  year={2020},
}

@inproceedings{simonyan2014deep,
  title={Deep inside convolutional networks: Visualising image classification models and saliency maps},
  author={Simonyan, Karen and Vedaldi, Andrea and Zisserman, Andrew},
  booktitle={ICLR Workshop},
  year={2014}
}

@article{samek2021explaining,
  title={Explaining deep neural networks and beyond: A review of methods and applications},
  author={Samek, Wojciech and Montavon, Gr{\'e}goire and Lapuschkin, Sebastian and Anders, Christopher J and M{\"u}ller, Klaus-Robert},
  journal={Proceedings of the IEEE},
  year={2021},
}

@article{samek2016evaluating,
  title={Evaluating the visualization of what a deep neural network has learned},
  author={Samek, Wojciech and Binder, Alexander and Montavon, Gr{\'e}goire and Lapuschkin, Sebastian and M{\"u}ller, Klaus-Robert},
  journal={TNNLS},
  year={2016},
}

@inproceedings{dabkowski2017real,
  title={Real time image saliency for black box classifiers},
  author={Dabkowski, Piotr and Gal, Yarin},
  booktitle={NeurIPS},
  year={2017}
}

@inproceedings{deyoung-etal-2020-eraser,
    title = "{ERASER}: {A} Benchmark to Evaluate Rationalized {NLP} Models",
    author = "DeYoung, Jay  and
      Jain, Sarthak  and
      Rajani, Nazneen Fatema  and
      Lehman, Eric  and
      Xiong, Caiming  and
      Socher, Richard  and
      Wallace, Byron C.",
    booktitle = {ACL},
    year = "2020",
}

@inproceedings{ismail2020benchmarking,
  title={Benchmarking Deep Learning Interpretability in Time Series Predictions},
  author={Ismail, Aya Abdelsalam and Gunady, Mohamed and Bravo, H{\'e}ctor Corrada and Feizi, Soheil},
  booktitle={NeurIPS},
  year={2020}
}

@article{bach2015pixel,
  title={On pixel-wise explanations for non-linear classifier decisions by layer-wise relevance propagation},
  author={Bach, Sebastian and Binder, Alexander and Montavon, Gr{\'e}goire and Klauschen, Frederick and M{\"u}ller, Klaus-Robert and Samek, Wojciech},
  journal={PloS one},
  volume={10},
  number={7},
  pages={e0130140},
  year={2015},
  publisher={Public Library of Science}
}

@inproceedings{fong2017interpretable,
  title={Interpretable explanations of black boxes by meaningful perturbation},
  author={Fong, Ruth C and Vedaldi, Andrea},
  booktitle={ICCV},
  pages={3429--3437},
  year={2017}
}

@article{krizhevsky2009learning,
  title={Learning multiple layers of features from tiny images},
  author={Krizhevsky, Alex and others},
  year={2009},
  publisher={Citeseer}
}

@article{netzer2011reading,
  title={Reading digits in natural images with unsupervised feature learning},
  author={Netzer, Yuval and Wang, Tao and Coates, Adam and Bissacco, Alessandro and Wu, Bo and Ng, Andrew Y},
  year={2011}
}

@techreport{WelinderEtal2010,
	Author = {P. Welinder and S. Branson and T. Mita and C. Wah and F. Schroff and S. Belongie and P. Perona},
	Institution = {California Institute of Technology},
	Number = {CNS-TR-2010-001},
	Title = {{Caltech-UCSD Birds 200}},
	Year = {2010}
}

@inproceedings{
    ancona2018towards,
    title={Towards better understanding of gradient-based attribution methods for Deep Neural Networks},
    author={Marco Ancona and Enea Ceolini and Cengiz Öztireli and Markus Gross},
    booktitle={ICLR},
    year={2018},
    url={https://openreview.net/forum?id=Sy21R9JAW},
}

@inproceedings{NIPS2017_8a20a862,
 author = {Lundberg, Scott M and Lee, Su-In},
 booktitle = {NeurIPS},
 editor = {I. Guyon and U. V. Luxburg and S. Bengio and H. Wallach and R. Fergus and S. Vishwanathan and R. Garnett},
 pages = {},
 publisher = {Curran Associates, Inc.},
 title = {A Unified Approach to Interpreting Model Predictions},
 url = {https://proceedings.neurips.cc/paper/2017/file/8a20a8621978632d76c43dfd28b67767-Paper.pdf},
 volume = {30},
 year = {2017}
}

@article{JMLR:v11:baehrens10a,
  author  = {David Baehrens and Timon Schroeter and Stefan Harmeling and Motoaki Kawanabe and Katja Hansen and Klaus-Robert M{{\"u}}ller},
  title   = {How to Explain Individual Classification Decisions},
  journal = {JMLR},
  year    = {2010},
  volume  = {11},
  number  = {61},
  pages   = {1803-1831},
  url     = {http://jmlr.org/papers/v11/baehrens10a.html}
}

@inproceedings{he2016deep,
  title={Deep residual learning for image recognition},
  author={He, Kaiming and Zhang, Xiangyu and Ren, Shaoqing and Sun, Jian},
  booktitle={CVPR},
  pages={770--778},
  year={2016}
}

@inproceedings{adebayo2018local,
  title={Local explanation methods for deep neural networks lack sensitivity to parameter values},
  author={Adebayo, Julius and Gilmer, Justin and Goodfellow, Ian and Kim, Been},
  booktitle={ICLR Workshop},
  year={2018}
}

@inproceedings{deng2009imagenet,
  title={Imagenet: A large-scale hierarchical image database},
  author={Deng, Jia and Dong, Wei and Socher, Richard and Li, Li-Jia and Li, Kai and Fei-Fei, Li},
  booktitle={CVPR},
  year={2009},
}

@inproceedings{loshchilov2016sgdr,
  title={SGDR: Stochastic gradient descent with warm restarts},
  author={Loshchilov, Ilya and Hutter, Frank},
  booktitle={ICLR},
  year={2017}
}

@article{virtanen2020scipy,
  title={SciPy 1.0: fundamental algorithms for scientific computing in Python},
  author={Virtanen, Pauli and Gommers, Ralf and Oliphant, Travis E and Haberland, Matt and Reddy, Tyler and Cournapeau, David and Burovski, Evgeni and Peterson, Pearu and Weckesser, Warren and Bright, Jonathan and others},
  journal={Nature methods},
  volume={17},
  number={3},
  pages={261--272},
  year={2020},
  publisher={Nature Publishing Group}
}

@inproceedings{khakzar2021neural,
  title={Neural Response Interpretation through the Lens of Critical Pathways},
  author={Khakzar, Ashkan and Baselizadeh, Soroosh and Khanduja, Saurabh and Rupprecht, Christian and Kim, Seong Tae and Navab, Nassir},
  booktitle={CVPR},
  pages={13528--13538},
  year={2021}
}

@inproceedings{hartley2021swag,
  title={Swag: Superpixels weighted by average gradients for explanations of cnns},
  author={Hartley, Thomas and Sidorov, Kirill and Willis, Christopher and Marshall, David},
  booktitle={WACV},
  pages={423--432},
  year={2021}
}

@inproceedings{chefer2021transformer,
  title={Transformer interpretability beyond attention visualization},
  author={Chefer, Hila and Gur, Shir and Wolf, Lior},
  booktitle={CVPR},
  pages={782--791},
  year={2021}
}

@inproceedings{zhang-etal-2021-sample,
    title = "On Sample Based Explanation Methods for {NLP}: Faithfulness, Efficiency and Semantic Evaluation",
    author = "Zhang, Wei  and
      Huang, Ziming  and
      Zhu, Yada  and
      Ye, Guangnan  and
      Cui, Xiaodong  and
      Zhang, Fan",
    booktitle = "IJCNLP",
    year = "2021",
}

@article{doshi2017towards,
  title={Towards a rigorous science of interpretable machine learning},
  author={Doshi-Velez, Finale and Kim, Been},
  journal={arXiv preprint arXiv:1702.08608},
  year={2017}
}

@article{meng2022interpretability,
  title={Interpretability and fairness evaluation of deep learning models on MIMIC-IV dataset},
  author={Meng, Chuizheng and Trinh, Loc and Xu, Nan and Enouen, James and Liu, Yan},
  journal={Scientific Reports},
  volume={12},
  number={1},
  pages={1--28},
  year={2022},
  publisher={Nature Publishing Group}
}

@article{zhou2021evaluating,
  title={Evaluating the quality of machine learning explanations: A survey on methods and metrics},
  author={Zhou, Jianlong and Gandomi, Amir H and Chen, Fang and Holzinger, Andreas},
  journal={Electronics},
  volume={10},
  number={5},
  pages={593},
  year={2021},
  publisher={Multidisciplinary Digital Publishing Institute}
}

@article{ismail2021improving,
  title={Improving Deep Learning Interpretability by Saliency Guided Training},
  author={Ismail, Aya Abdelsalam and Corrada Bravo, Hector and Feizi, Soheil},
  journal={NeurIPS},
  volume={34},
  year={2021}
}

@inproceedings{rong2022consistent,
  title={A Consistent and Efficient Evaluation Strategy for Attribution Methods},
  author={Rong, Yao and Leemann, Tobias and Borisov, Vadim and Kasneci, Gjergji and Kasneci, Enkelejda},
  booktitle={ICML},
  pages={18770--18795},
  year={2022},
  organization={PMLR}
}

@article{hellman1970probability,
  title={Probability of error, equivocation, and the Chernoff bound},
  author={Hellman, Martin and Raviv, Josef},
  journal={IEEE Transactions on Information Theory},
  volume={16},
  number={4},
  pages={368--372},
  year={1970},
  publisher={IEEE}
}

@article{funke2022z,
  title={Zorro: Valid, sparse, and stable explanations in graph neural networks},
  author={Funke, Thorben and Khosla, Megha and Rathee, Mandeep and Anand, Avishek},
  journal={IEEE Transactions on Knowledge and Data Engineering},
  year={2022},
  publisher={IEEE}
}

@article{nauta2022anecdotal,
  title={From anecdotal evidence to quantitative evaluation methods: A systematic review on evaluating explainable ai},
  author={Nauta, Meike and Trienes, Jan and Pathak, Shreyasi and Nguyen, Elisa and Peters, Michelle and Schmitt, Yasmin and Schl{\"o}tterer, J{\"o}rg and van Keulen, Maurice and Seifert, Christin},
  journal={arXiv preprint arXiv:2201.08164},
  year={2022}
}

@article{yeh2019fidelity,
  title={On the (in)fidelity and sensitivity of explanations},
  author={Yeh, Chih-Kuan and Hsieh, Cheng-Yu and Suggala, Arun and Inouye, David I and Ravikumar, Pradeep K},
  journal={NeurIPS},
  volume={32},
  year={2019}
}

@misc{2020mmclassification,
    title={OpenMMLab's Image Classification Toolbox and Benchmark},
    author={MMClassification Contributors},
    howpublished = {\url{https://github.com/open-mmlab/mmclassification}},
    year={2020}
}

@article{liu2022rethinking,
  title={Rethinking Efficiency and Redundancy in Training Large-scale Graphs},
  author={Liu, Xin and Xiong, Xunbin and Yan, Mingyu and Xue, Runzhen and Pan, Shirui and Ye, Xiaochun and Fan, Dongrui},
  journal={arXiv preprint arXiv:2209.00800},
  year={2022}
}

@article{radovic2017minimum,
  title={Minimum redundancy maximum relevance feature selection approach for temporal gene expression data},
  author={Radovic, Milos and Ghalwash, Mohamed and Filipovic, Nenad and Obradovic, Zoran},
  journal={BMC bioinformatics},
  volume={18},
  number={1},
  pages={1--14},
  year={2017},
  publisher={BioMed Central}
}

@article{park2024geometric,
  title={Geometric remove-and-retrain (goar): Coordinate-invariant explainable ai assessment},
  author={Park, Yong-Hyun and Seo, Junghoon and Park, Bomseok and Lee, Seongsu and Jo, Junghyo},
  journal={NeurIPS Workshop},
  year={2023}
}

@inproceedings{hong2025comprehensive,
  title={Comprehensive Information Bottleneck for Unveiling Universal Attribution to Interpret Vision Transformers},
  author={Hong, Jung-Ho and Kim, Ho-Joong and Jeon, Kyu-Sung and Lee, Seong-Whan},
  booktitle={Proceedings of the Computer Vision and Pattern Recognition Conference},
  pages={25166--25175},
  year={2025}
}

@inproceedings{duan2024evaluation,
  title={On the evaluation consistency of attribution-based explanations},
  author={Duan, Jiarui and Li, Haoling and Zhang, Haofei and Jiang, Hao and Xue, Mengqi and Sun, Li and Song, Mingli and Song, Jie},
  booktitle={European Conference on Computer Vision},
  pages={206--224},
  year={2024},
  organization={Springer}
}

@inproceedings{
zheng2025ffidelity,
title={F-Fidelity: A Robust Framework for Faithfulness Evaluation of Explainable {AI}},
author={Xu Zheng and Farhad Shirani and Zhuomin Chen and Chaohao Lin and Wei Cheng and Wenbo Guo and Dongsheng Luo},
booktitle={The Thirteenth International Conference on Learning Representations},
year={2025},
url={https://openreview.net/forum?id=X0r4BN50Dv}
}

@article{bora2025fries,
  title={FRIES: Framework for inconsistency estimation of saliency metrics},
  author={Bora, Revoti Prasad and Terh{\"o}rst, Philipp and Veldhuis, Raymond and Ramachandra, Raghavendra and Raja, Kiran},
  journal={Pattern Recognition},
  pages={112136},
  year={2025},
  publisher={Elsevier}
}

@article{canha2025functionally,
  title={A Functionally-Grounded Benchmark Framework for XAI Methods: Insights and Foundations from a Systematic Literature Review},
  author={Canha, Dulce and Kubler, Sylvain and Fr{\"a}mling, Kary and Fagherazzi, Guy},
  journal={ACM Computing Surveys},
  year={2025},
  publisher={ACM New York, NY}
}

@article{sharkey2025open,
  title={Open problems in mechanistic interpretability},
  author={Sharkey, Lee and Chughtai, Bilal and Batson, Joshua and Lindsey, Jack and Wu, Jeff and Bushnaq, Lucius and Goldowsky-Dill, Nicholas and Heimersheim, Stefan and Ortega, Alejandro and Bloom, Joseph and others},
  journal={arXiv preprint arXiv:2501.16496},
  year={2025}
}

@inproceedings{wang2023interpretability,
title={Interpretability in the Wild: a Circuit for Indirect Object Identification in {GPT}-2 Small},
author={Kevin Ro Wang and Alexandre Variengien and Arthur Conmy and Buck Shlegeris and Jacob Steinhardt},
booktitle={The Eleventh International Conference on Learning Representations },
year={2023},
url={https://openreview.net/forum?id=NpsVSN6o4ul}
}

@article{bricken2023monosemanticity,
  title={Towards Monosemanticity: Decomposing Language Models With Dictionary Learning},
  author={Bricken, Trenton and Templeton, Adly and Batson, Joshua and Chen, Brian and Jermyn, Adam and Conerly, Tom and Turner, Nick and Anil, Cem and Denison, Carson and Askell, Amanda and others},
  journal={Transformer Circuits Thread},
  year={2023}
}

\newpage

\appendix

\section{Proof of Theorem~\ref{theorem:dpi}}
\label{app:proof_dpi}

\begin{proof}[Proof of Theorem~\ref{theorem:dpi}]
Fix any $x$.
Because $\widetilde{A}$ is generated from $(A,U)$ with $U$ independent of all other variables, we have the conditional Markov chain
\begin{equation}
  Z \;\to\; A \;\to\; \widetilde{A}
  \quad\text{given } X=x,
\end{equation}
i.e., $\widetilde{A}\perp Z \mid (A,X=x)$.
By the (conditional) data processing inequality,
\begin{equation}
  I(Z;\widetilde{A}\mid X=x) \;\le\; I(Z;A\mid X=x).
\end{equation}
Taking expectation over $X$ yields $I(Z;\widetilde{A}\mid X)\le I(Z;A\mid X)$.

For the second claim, given $X=x$ the random variable $\psi(X,\widetilde{A})$ is a function of $\widetilde{A}$ (and independent randomness internal to $\psi$).
Thus we have another Markov chain
\begin{equation}
  Z \;\to\; \widetilde{A} \;\to\; \psi(X,\widetilde{A})
  \quad\text{given } X=x,
\end{equation}
and DPI implies $I(Z;\psi(X,\widetilde{A})\mid X=x)\le I(Z;\widetilde{A}\mid X=x)$.
Averaging over $X$ proves the statement, and the cases $\widetilde{M}_t=h_t(\widetilde{A})$ and $\widetilde{X}_t^\prime=\phi(X,\widetilde{M}_t)$ follow by choosing $\psi$ appropriately.
\end{proof}

\section{Details of Experiment Setting}
\label{app:exp_details}
All models were trained for 50 epochs with batch size $b=128$ except for $b=32$ on CUB-200~\citep{WelinderEtal2010} on a single GTX1080Ti GPU. Each model was trained with SGD optimizer with momentum $\beta=0.9$ and weight decay $\lambda=1\times10^{-4}$ except for $\lambda=5\times10^{-4}$ on CIFAR-10.
To avoid convergence on bad local minima, cosine annealing scheduling~\citep{loshchilov2016sgdr} was used, starting from the initial learning rate $\alpha=0.1$ for CIFAR-10 and SHVN. For CUB-200, $\alpha=0.01$ was used. The learning rate was annealed to zero over 50 epochs. Especially in the case of CUB-200, we use ImageNet~\citep{deng2009imagenet} pre-trained model for fast convergence.
In our ROAD~\citep{rong2022consistent} experiments, we utilized MMClassification framework~\citep{2020mmclassification} to facilitate the implementation. Our training strategies followed MMClassification's ones for ResNet18 on CIFAR-10 and ResNet50 on CUB-200. In the case of SVHN, we adopted the same strategy as CIFAR-10.

Base estimators were implemented by PyTorch AutoGrad module. Similar to~\cite{hooker2019benchmark}, in Integrated Gradients, interval $k$ is set to 25, and the reference image is set to zeros. Also, for the ensemble methodologies, the sample size is set to $n=15$. Among all estimators, only Sobel edge detector is implemented by SciPy library~\citep{virtanen2020scipy}.
Finally, we post-process attribution map using Maximum filter with kernel size $=3$ and Gaussian filter function with standard deviation for Gaussian kernel $\sigma=1$ in SciPy library.

\clearpage
\onecolumn

\section{Additional Tables and Plots}
\label{app:plot}

\begin{figure}[H]
\centering
\begin{tabular}{c | c  c  c}\hline
 & \small \bf 10\% & \small \bf 30\% & \small \bf 50\% \\\hline
\small \bf \rotatebox[origin=l]{90}{CIFAR-10} & \subfigure[\small $R^2=0.84$]{\includegraphics[width=0.20\textwidth]{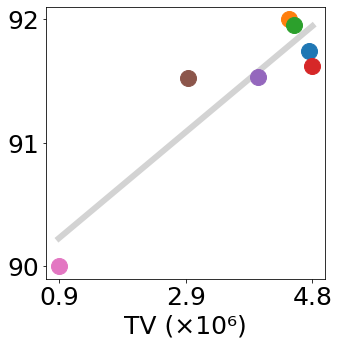}} &
\subfigure[$R^2=0.85$]{\includegraphics[width=0.20\textwidth]{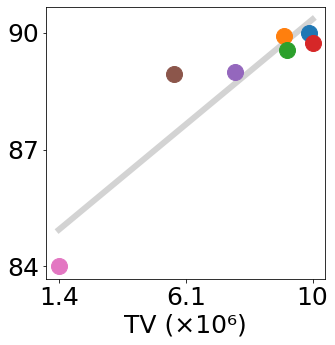}} &
\subfigure[$R^2=0.82$]{\includegraphics[width=0.20\textwidth]{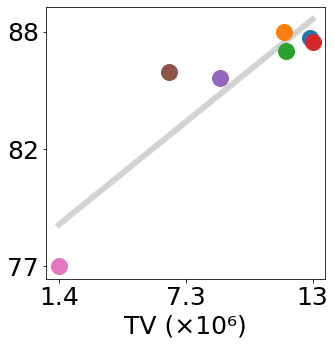}}\\
\small \bf \rotatebox[origin=l]{90}{SVHN} & \subfigure[$R^2=0.92$]{\includegraphics[width=0.20\textwidth]{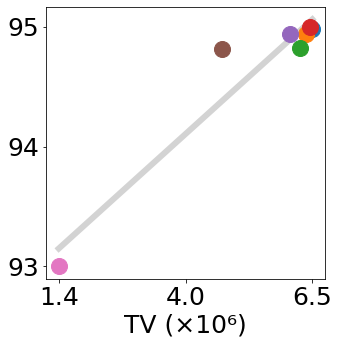}} &
\subfigure[$R^2=0.90$]{\includegraphics[width=0.20\textwidth]{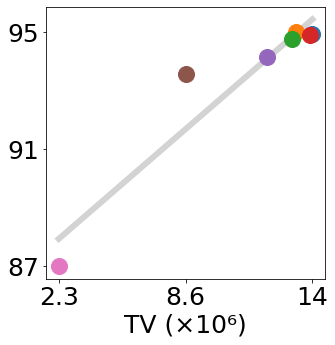}} &
\subfigure[$R^2=0.95$]{\includegraphics[width=0.20\textwidth]{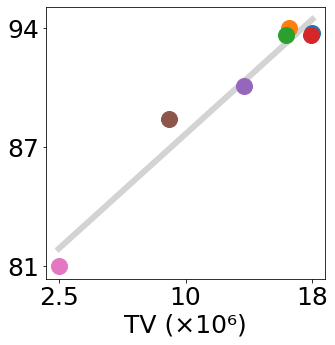}}\\
\small \bf \rotatebox[origin=l]{90}{CUB-200} & \subfigure[$R^2=0.91$]{\includegraphics[width=0.20\textwidth]{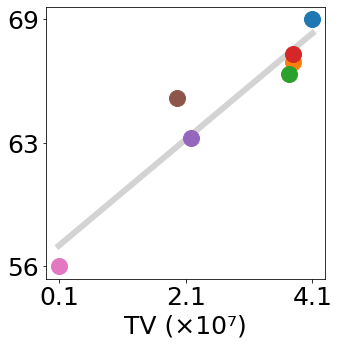}} &
\subfigure[$R^2=0.90$]{\includegraphics[width=0.20\textwidth]{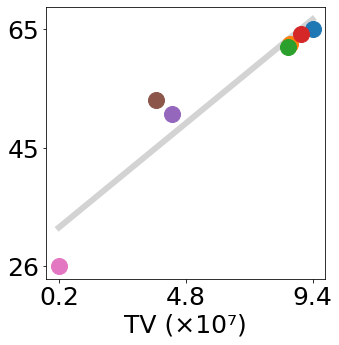}} &
\subfigure[$R^2=0.92$]{\includegraphics[width=0.20\textwidth]{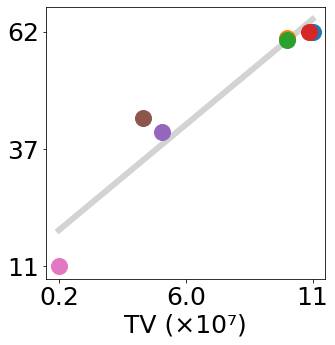}}
\end{tabular}\\
\subfigure{\includegraphics[width=0.55\textwidth]{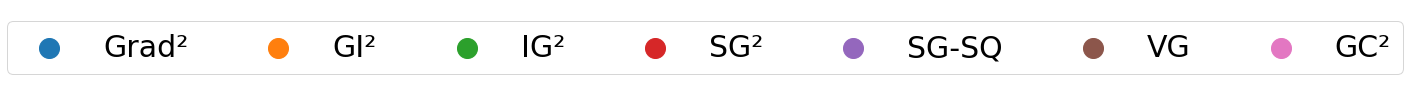}}
\caption{The relationship between model accuracy and the total variation of attribution masks in terms of attribution method. The fitted line and coefficient of determination from simple linear regression are also included. The y-axis represents the final test accuracy (\%).}
\label{fig:relation-sq}
\end{figure}
\twocolumn

\noindent\vbox to \textheight{\vfil\hsize=\columnwidth
\begin{figure}[H]
\centering
\begin{tabular}{c c c}
\subfigure[Grad$^2$]{\includegraphics[width=0.30\columnwidth]{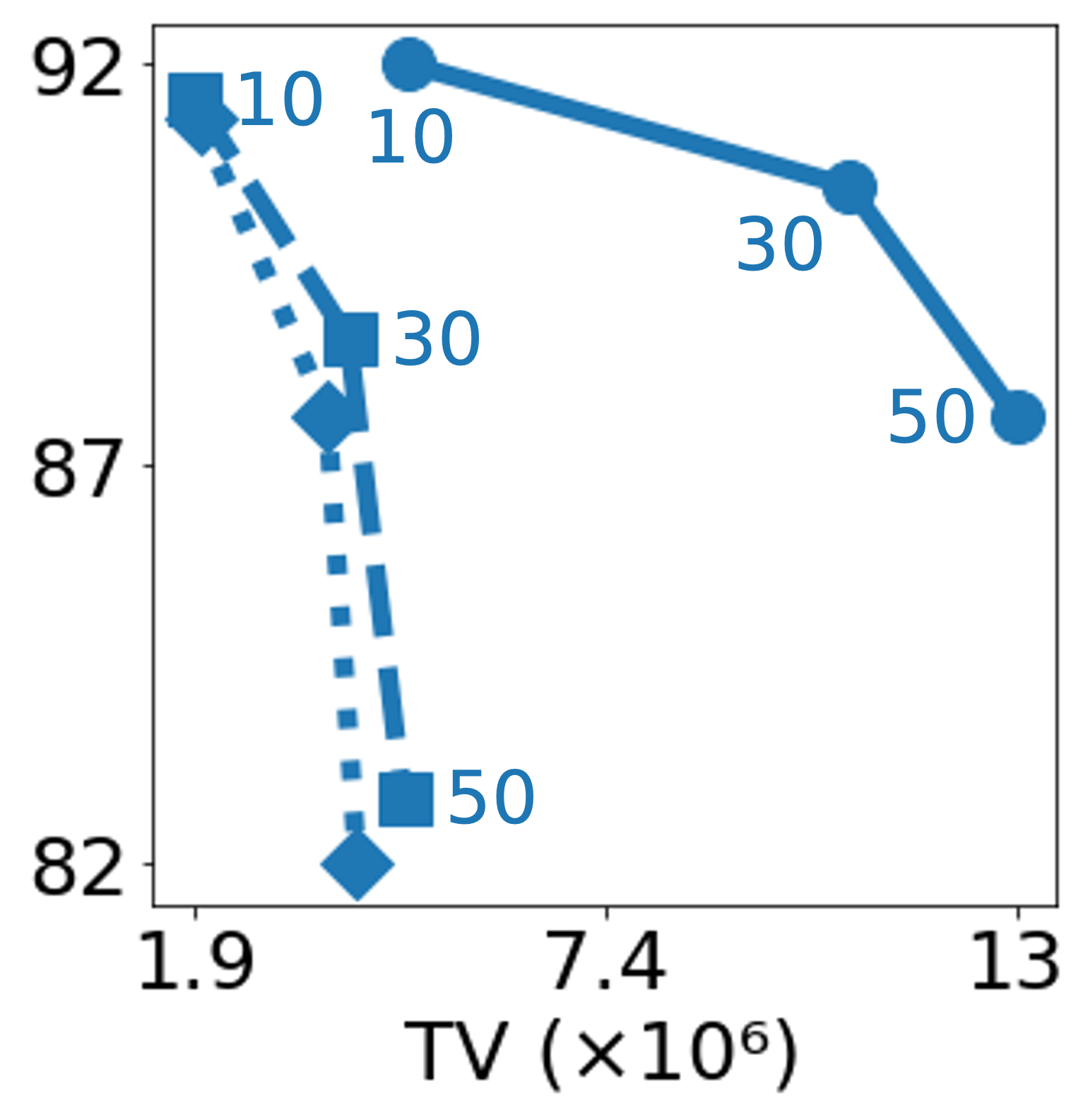}} &
\subfigure[GI$^2$]{\includegraphics[width=0.30\columnwidth]{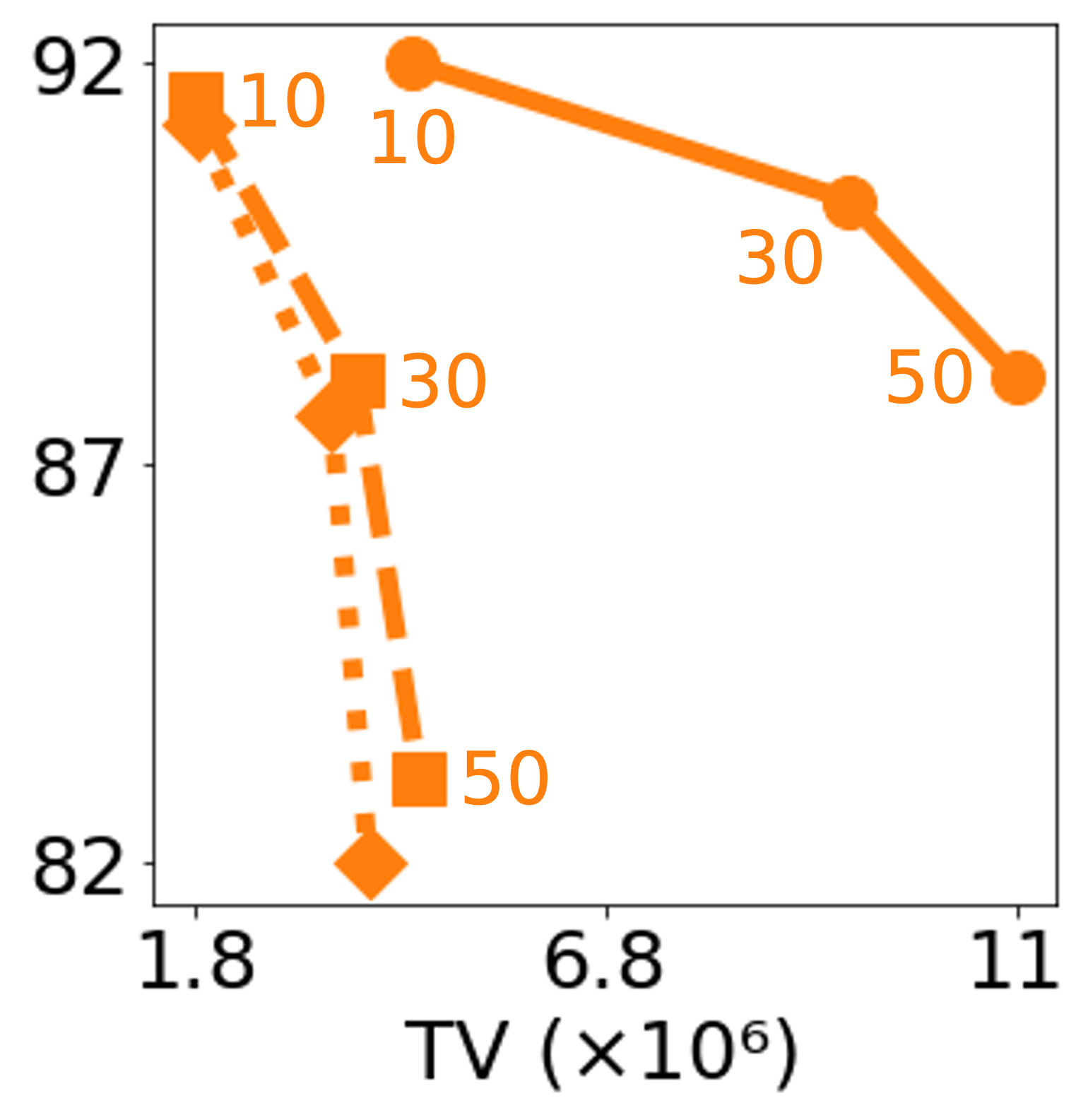}} &
\subfigure[IG$^2$]{\includegraphics[width=0.30\columnwidth]{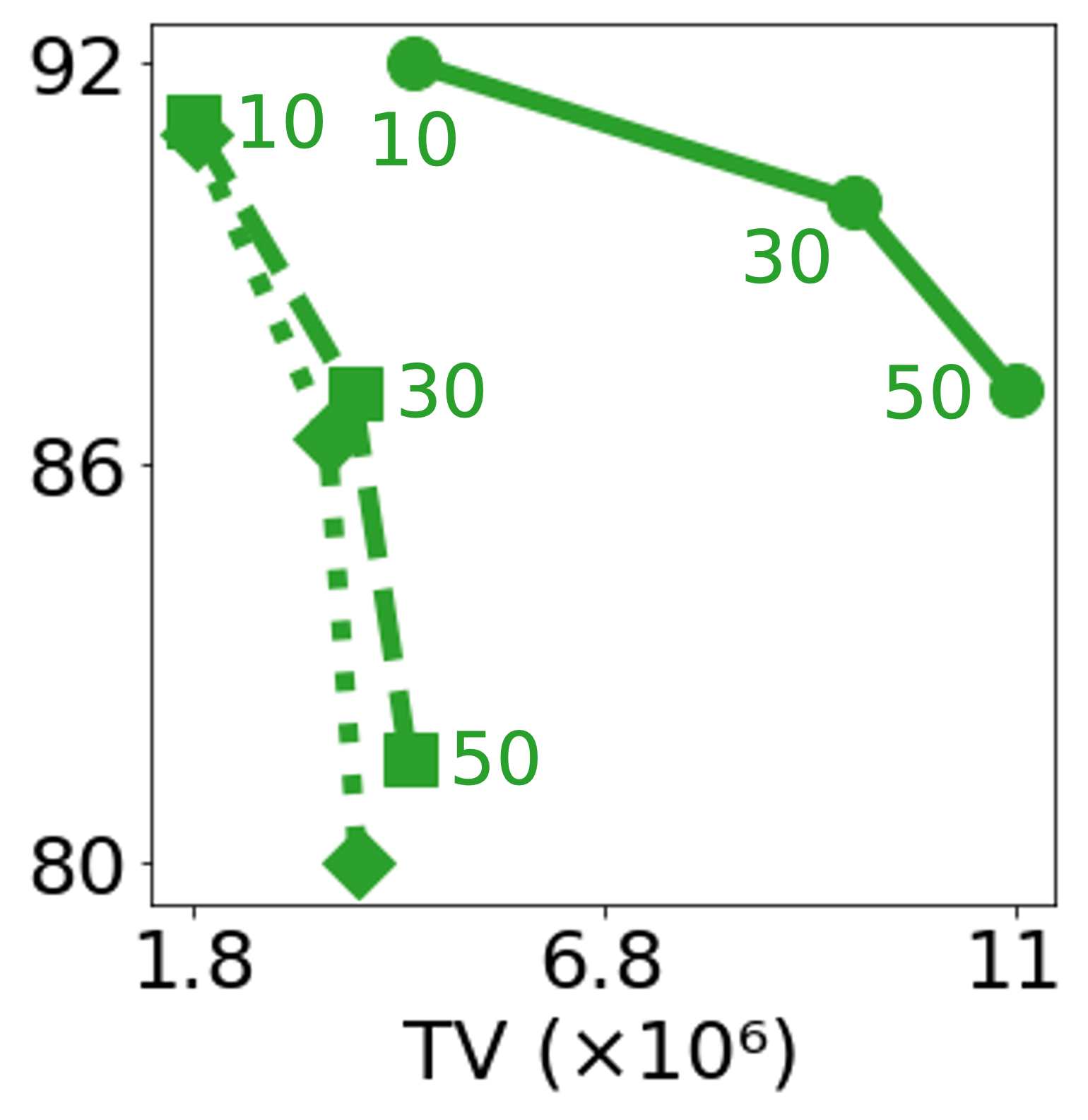}}\\
\subfigure[SG$^2$]{\includegraphics[width=0.30\columnwidth]{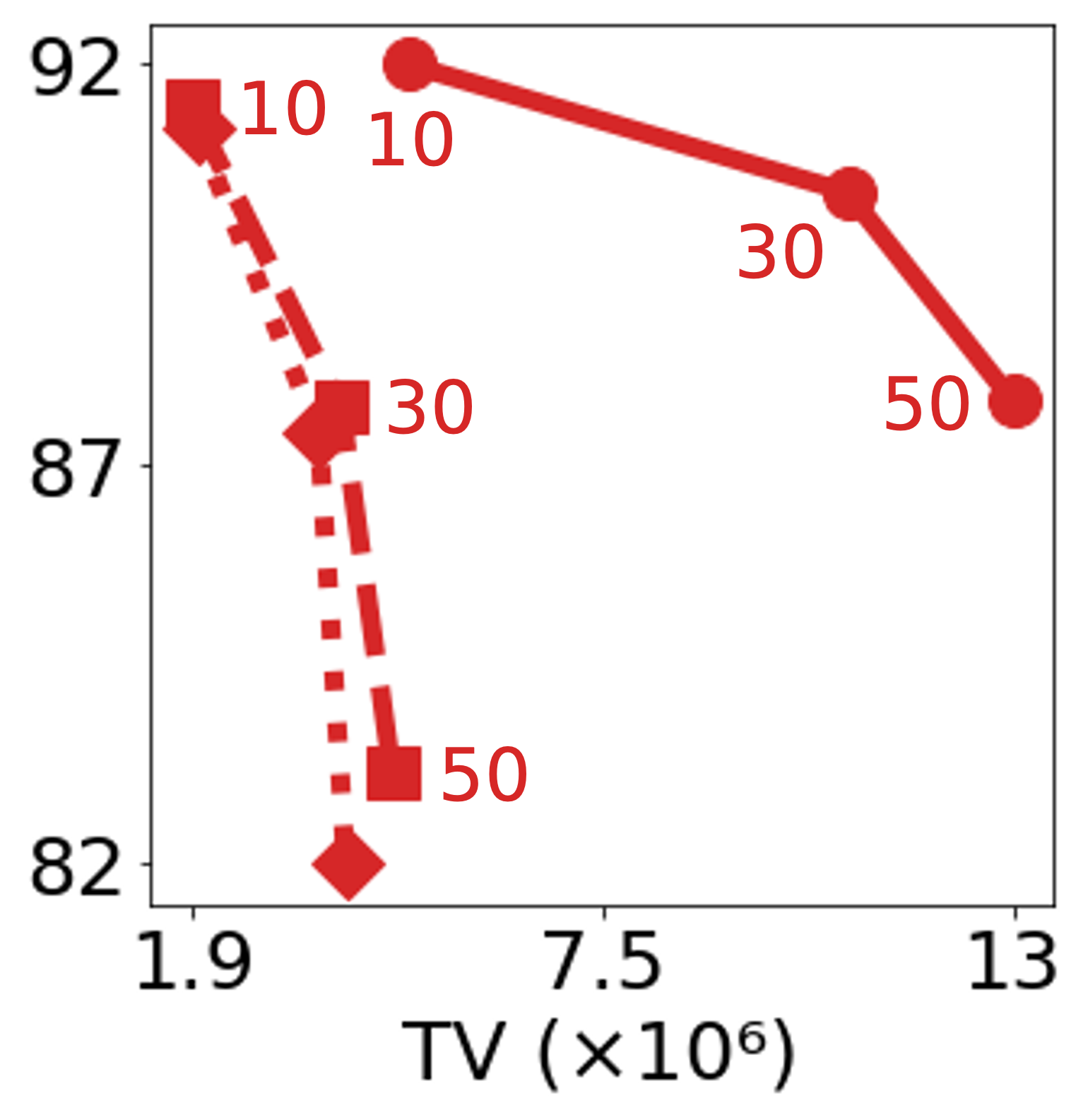}} &
\subfigure[SG-SQ]{\includegraphics[width=0.30\columnwidth]{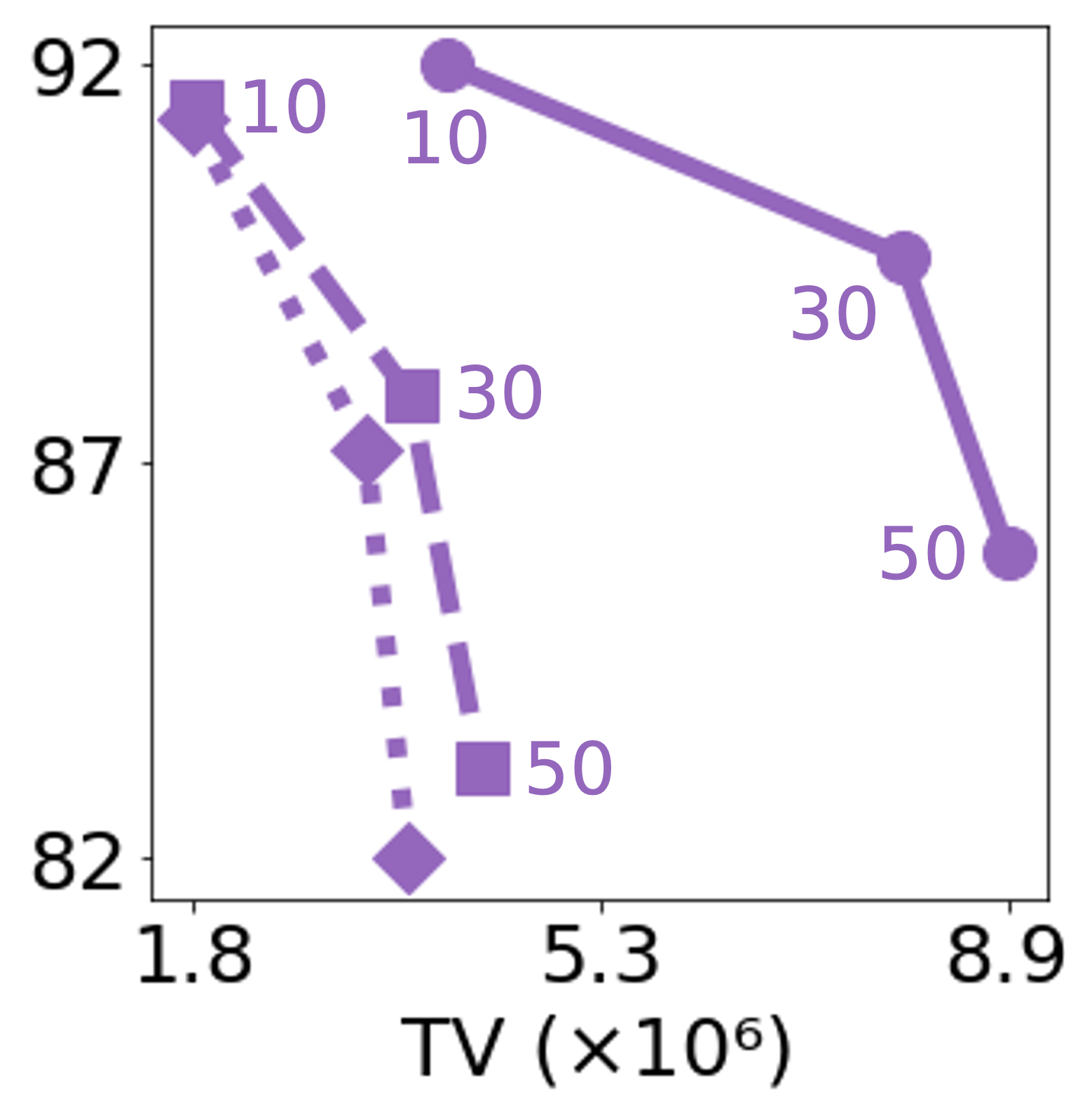}} &
\subfigure[VG]{\includegraphics[width=0.30\columnwidth]{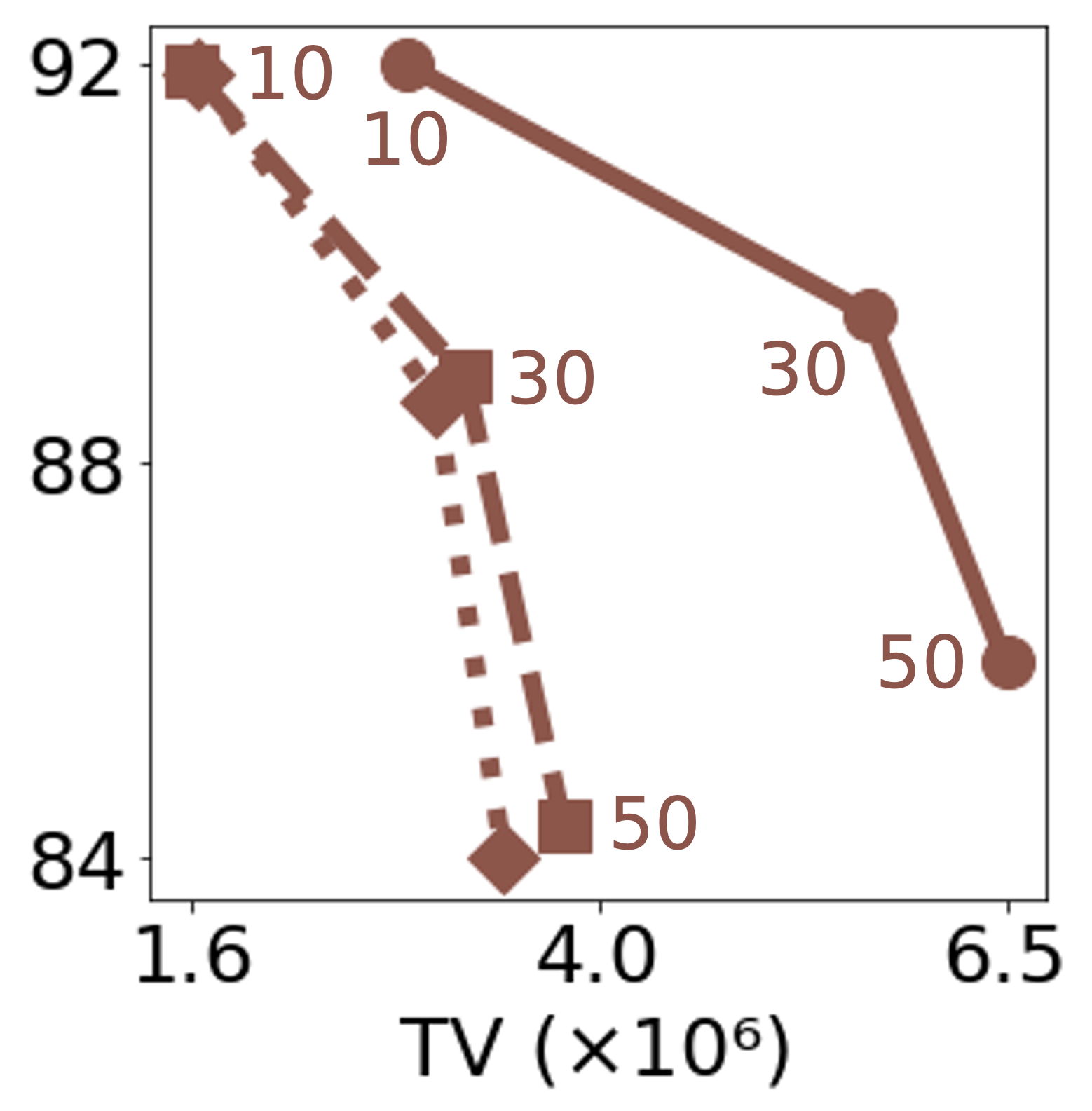}}\\
\subfigure[GC$^2$]{\includegraphics[width=0.30\columnwidth]{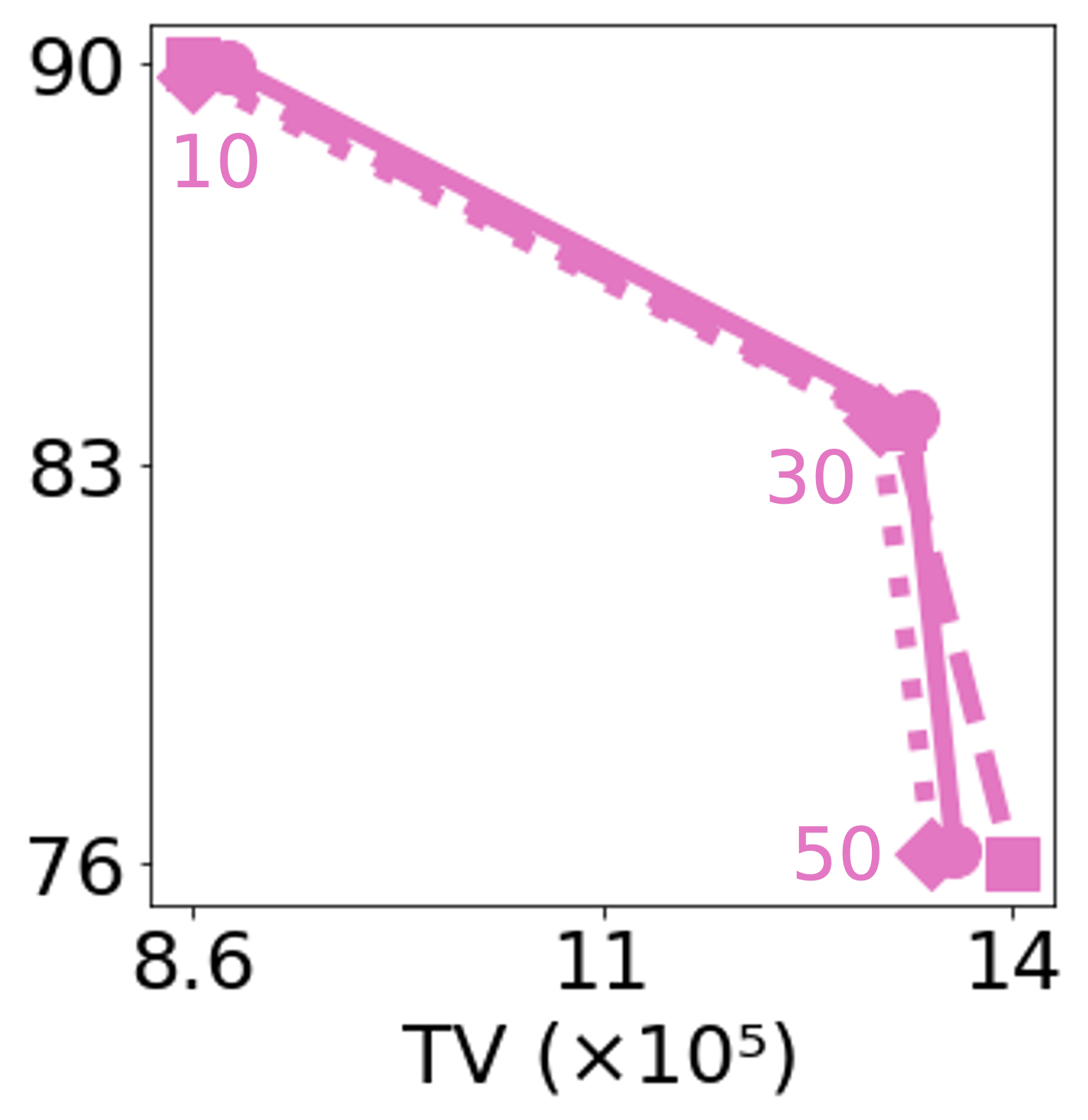}} &
\subfigure[Sobel$^2$]{\includegraphics[width=0.30\columnwidth]{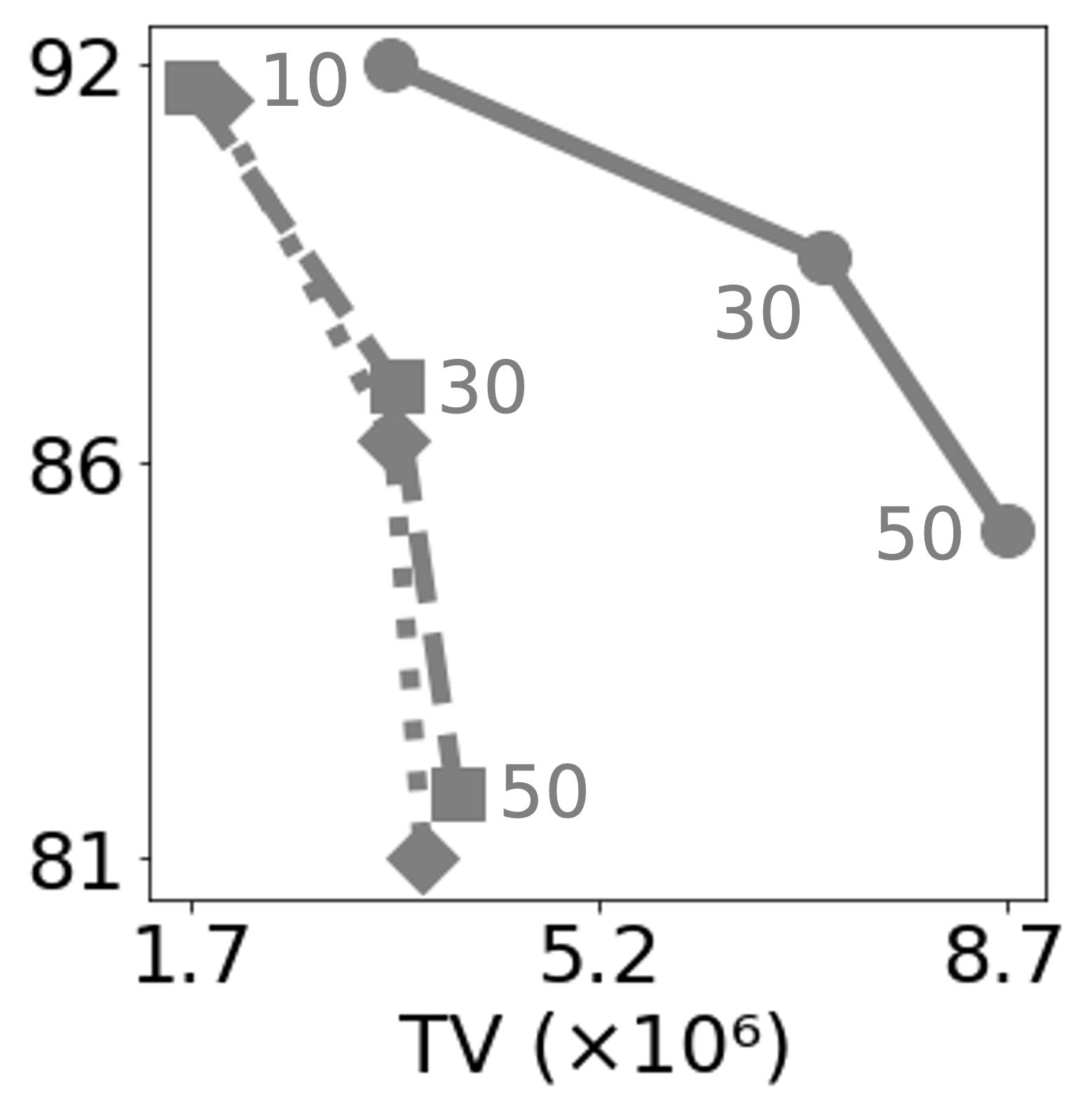}} &
\subfigure[Rand]{\includegraphics[width=0.30\columnwidth]{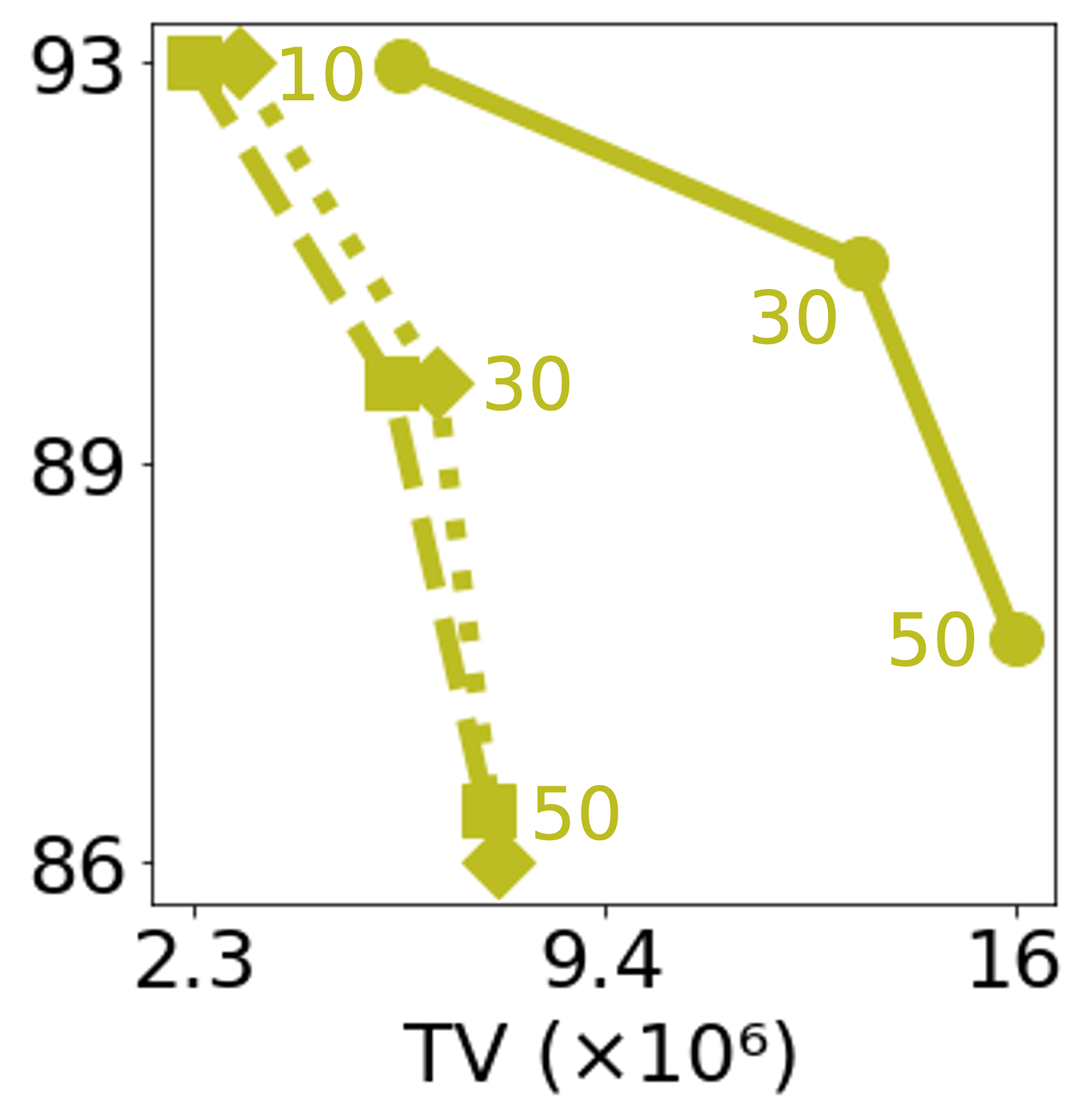}}
\end{tabular}\\
\subfigure{\includegraphics[width=0.7\columnwidth]{figure/1/legend.png}}
\caption{Effects of Gaussian filtering and max-pooling on total variation of attribution masks and model accuracy on CIFAR-10.}
\label{fig:tv-accuracy-cifar10}
\end{figure}
\vfil}
\newpage

\noindent\vbox to \textheight{\vfil\hsize=\columnwidth
\begin{figure}[H]
\centering
\begin{tabular}{c c c}
\subfigure[Grad$^2$]{\includegraphics[width=0.30\columnwidth]{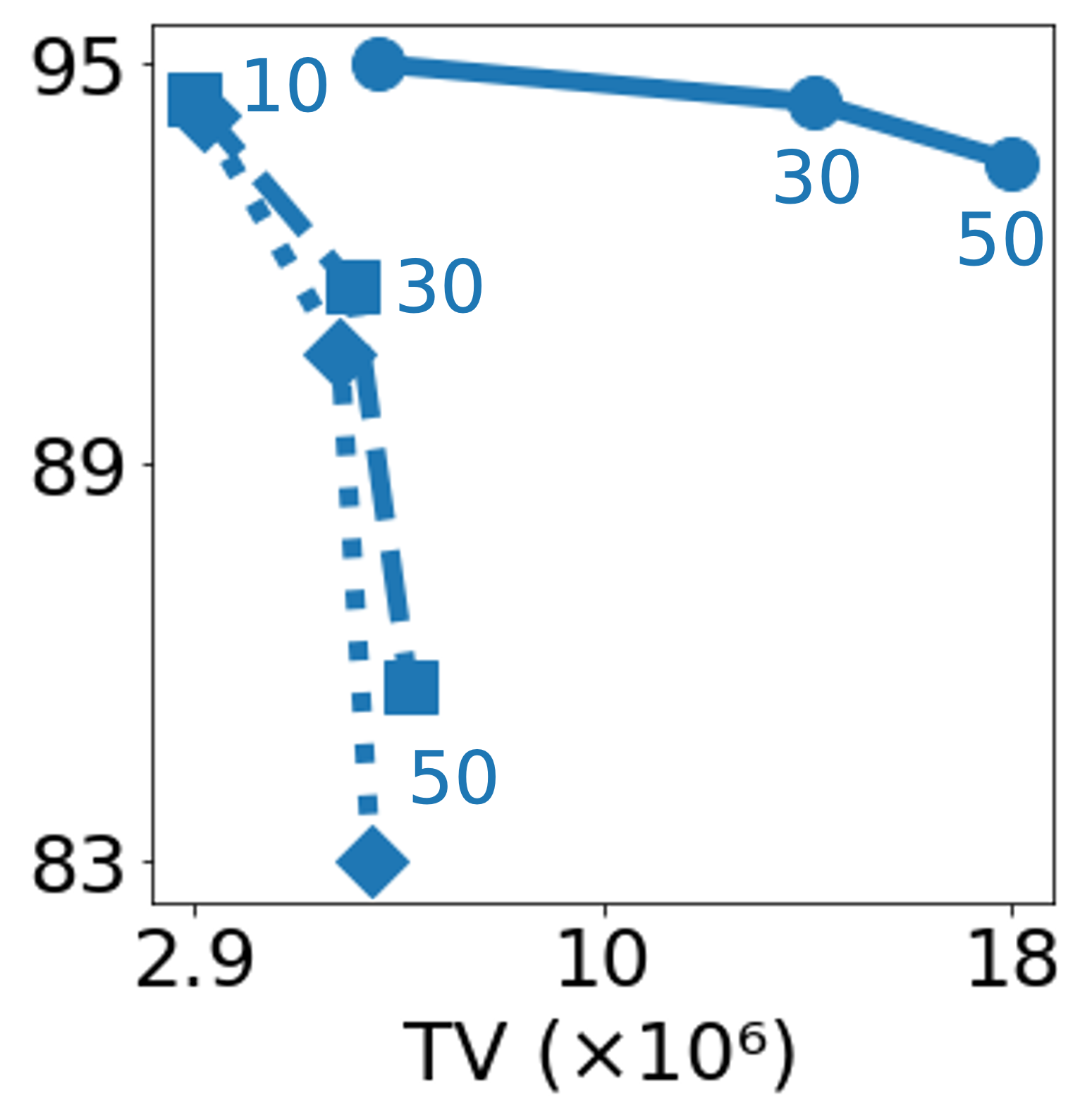}} &
\subfigure[GI$^2$]{\includegraphics[width=0.30\columnwidth]{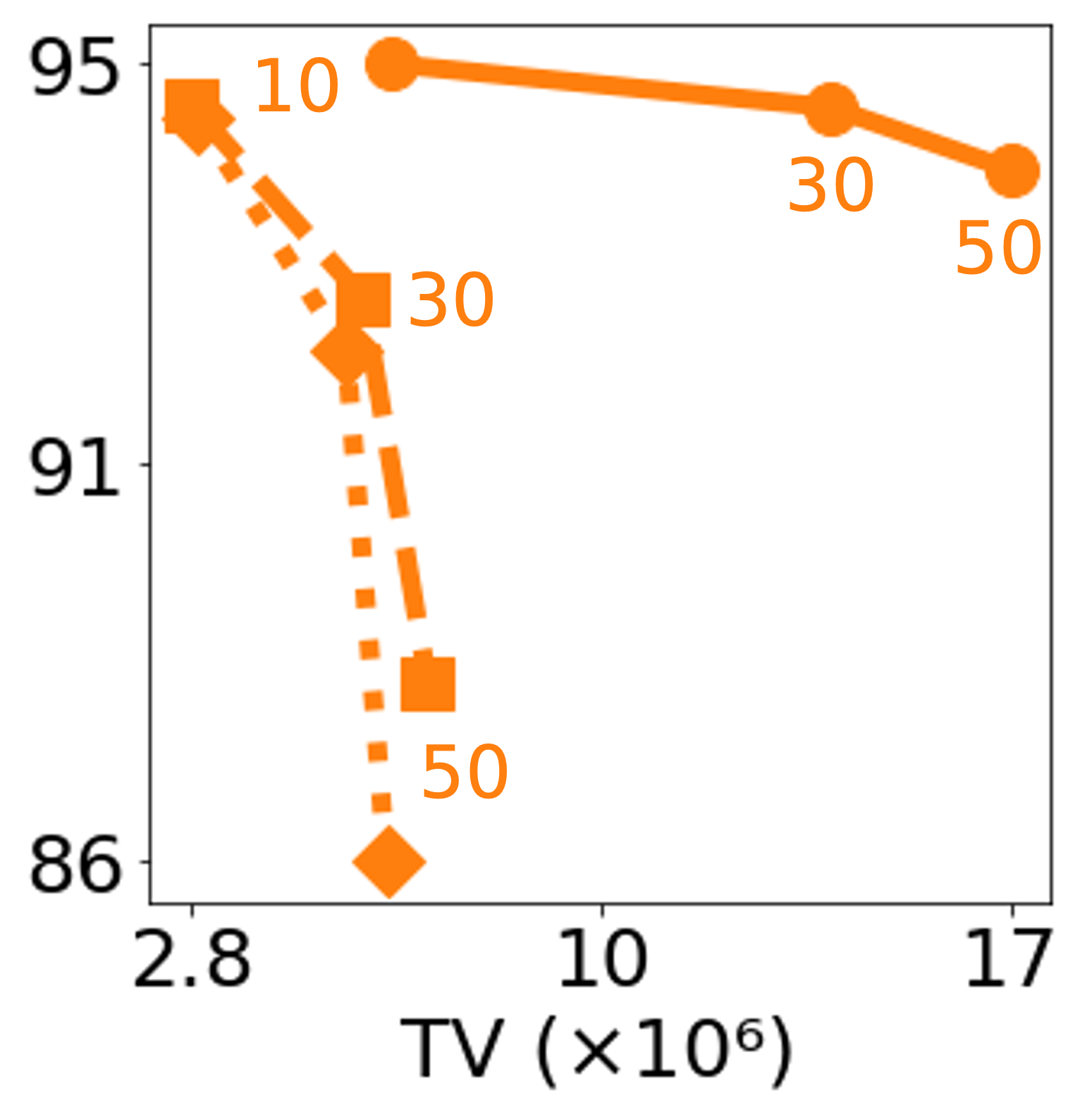}} &
\subfigure[IG$^2$]{\includegraphics[width=0.30\columnwidth]{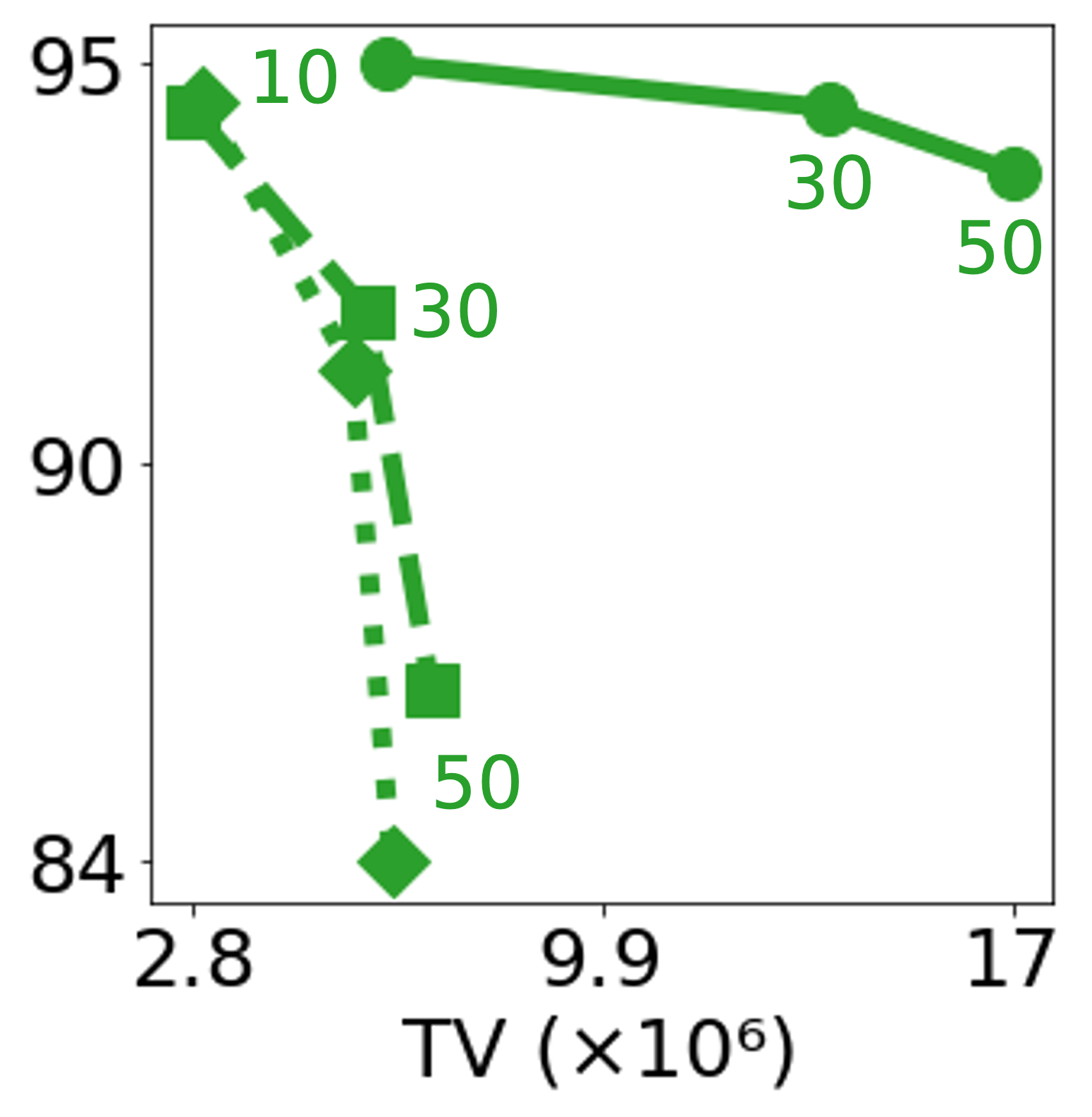}}\\
\subfigure[SG$^2$]{\includegraphics[width=0.30\columnwidth]{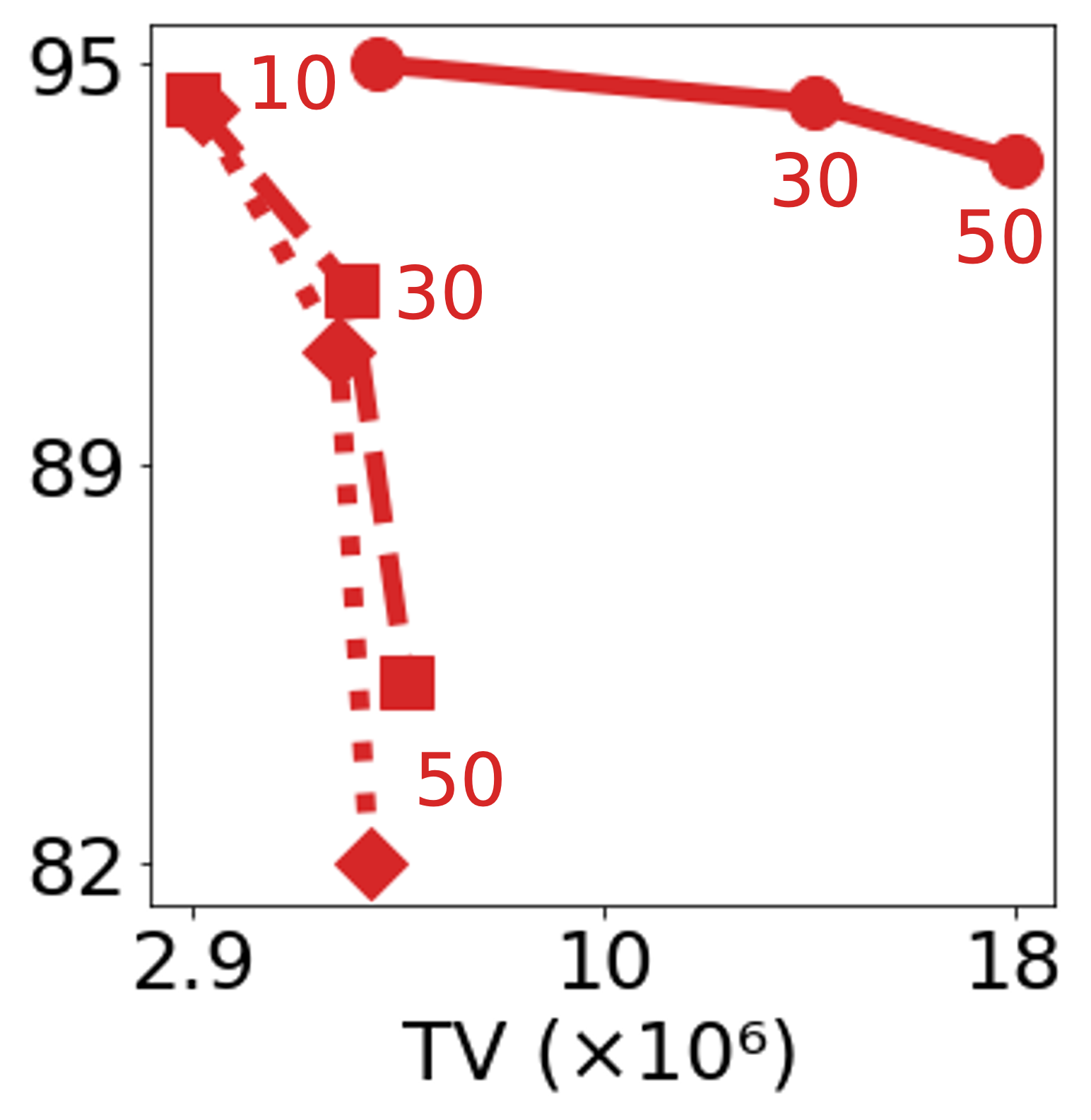}} &
\subfigure[SG-SQ]{\includegraphics[width=0.30\columnwidth]{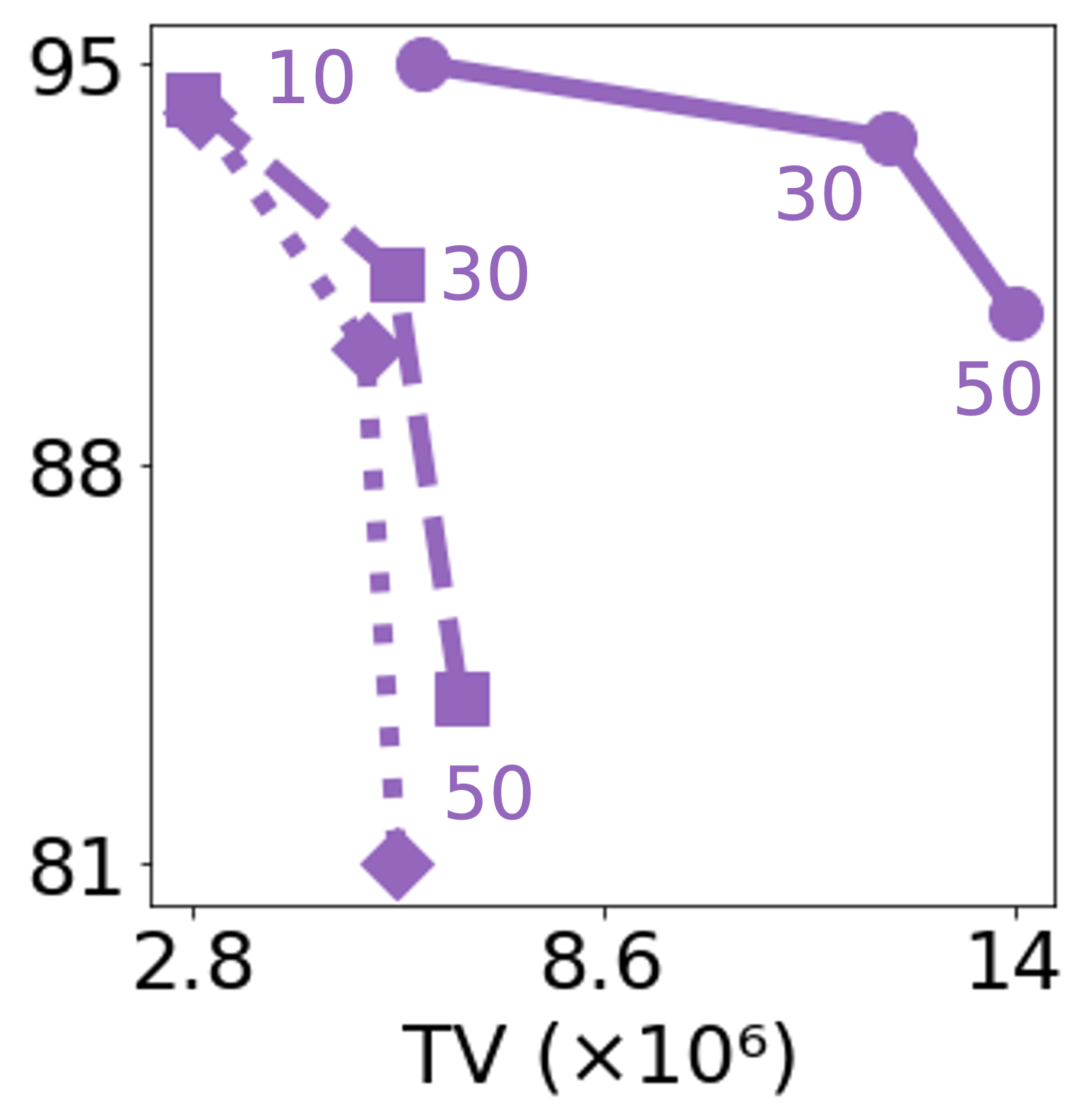}} &
\subfigure[VG]{\includegraphics[width=0.30\columnwidth]{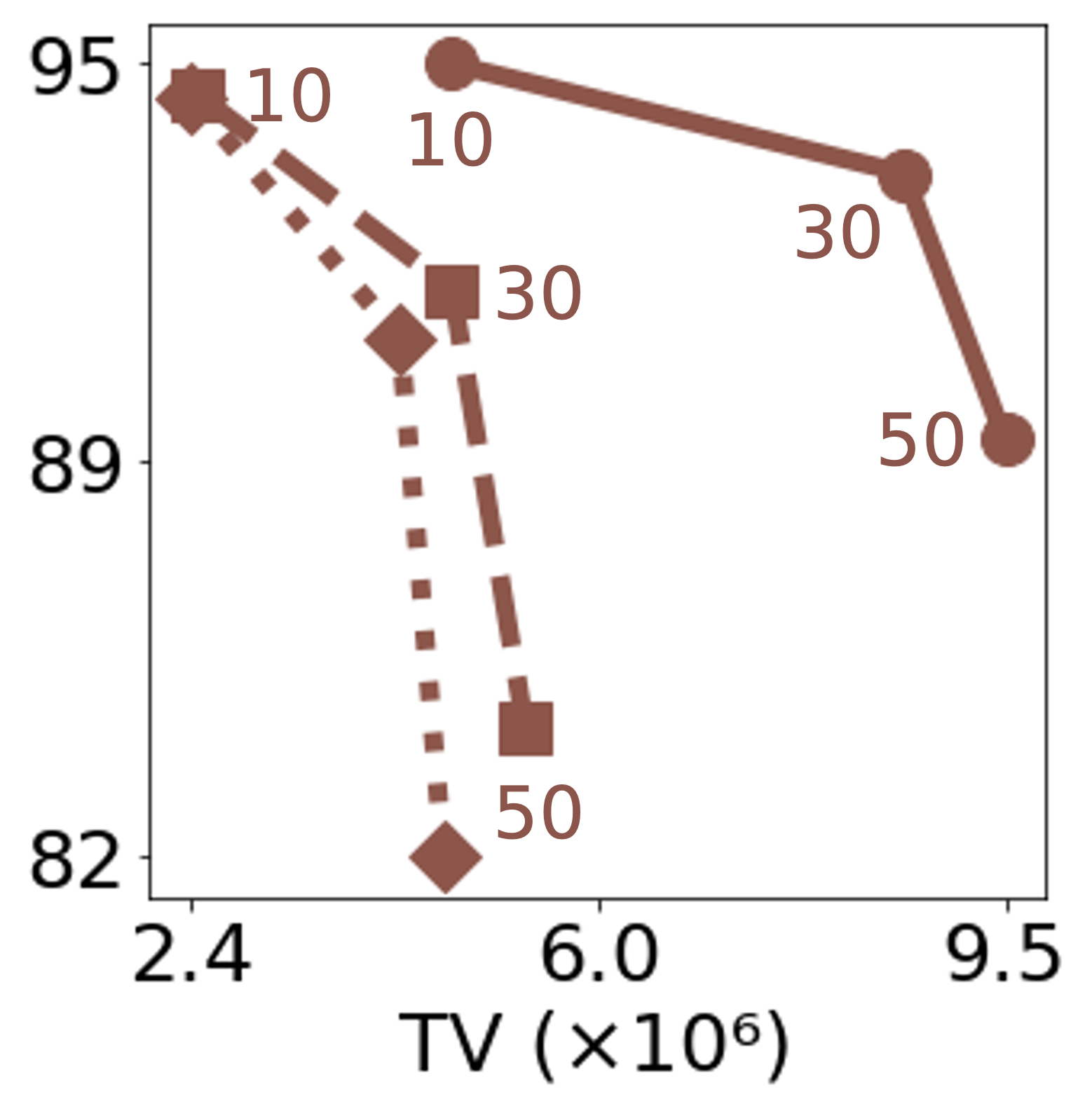}}\\
\subfigure[GC$^2$]{\includegraphics[width=0.30\columnwidth]{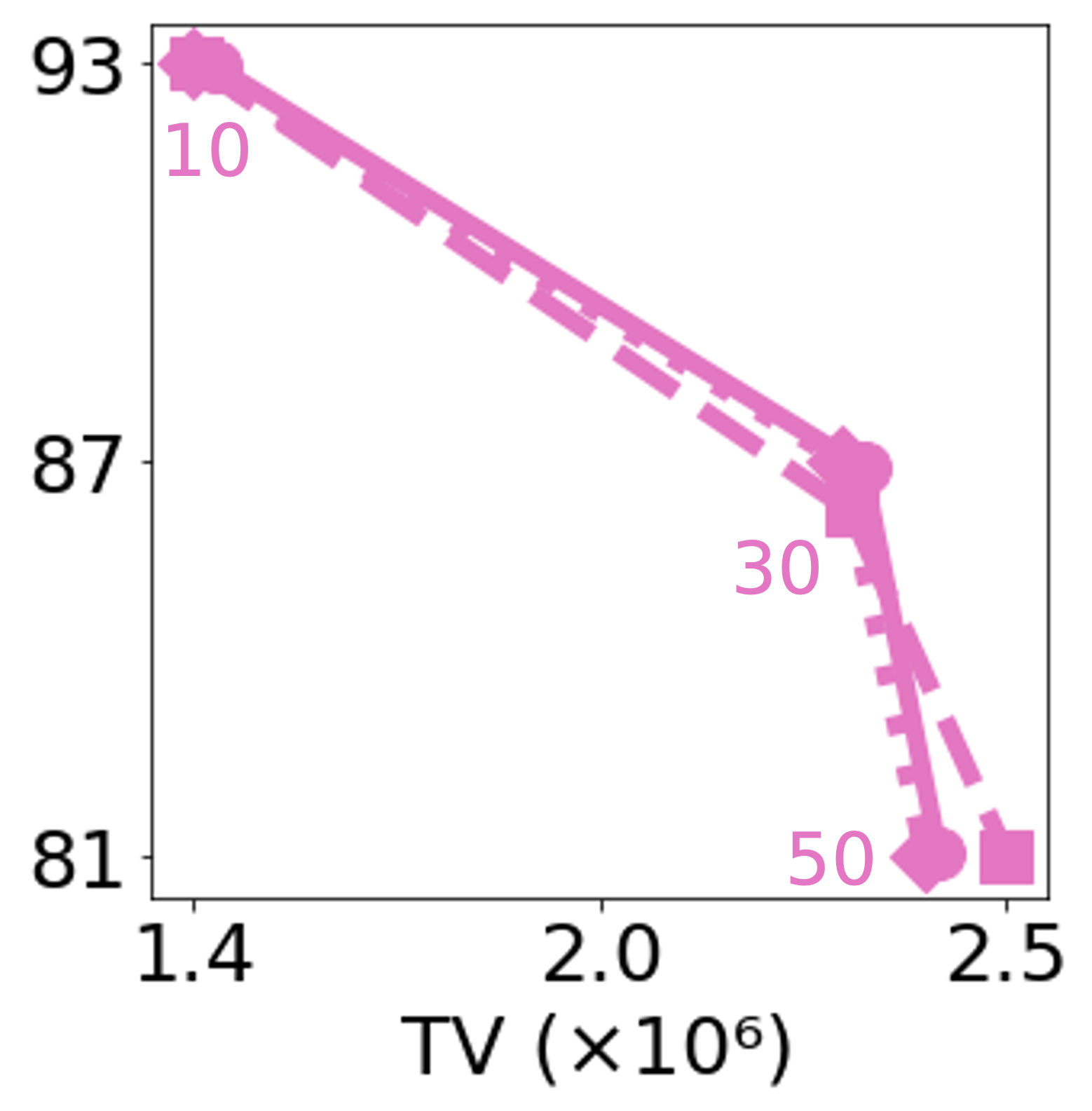}} &
\subfigure[Sobel$^2$]{\includegraphics[width=0.30\columnwidth]{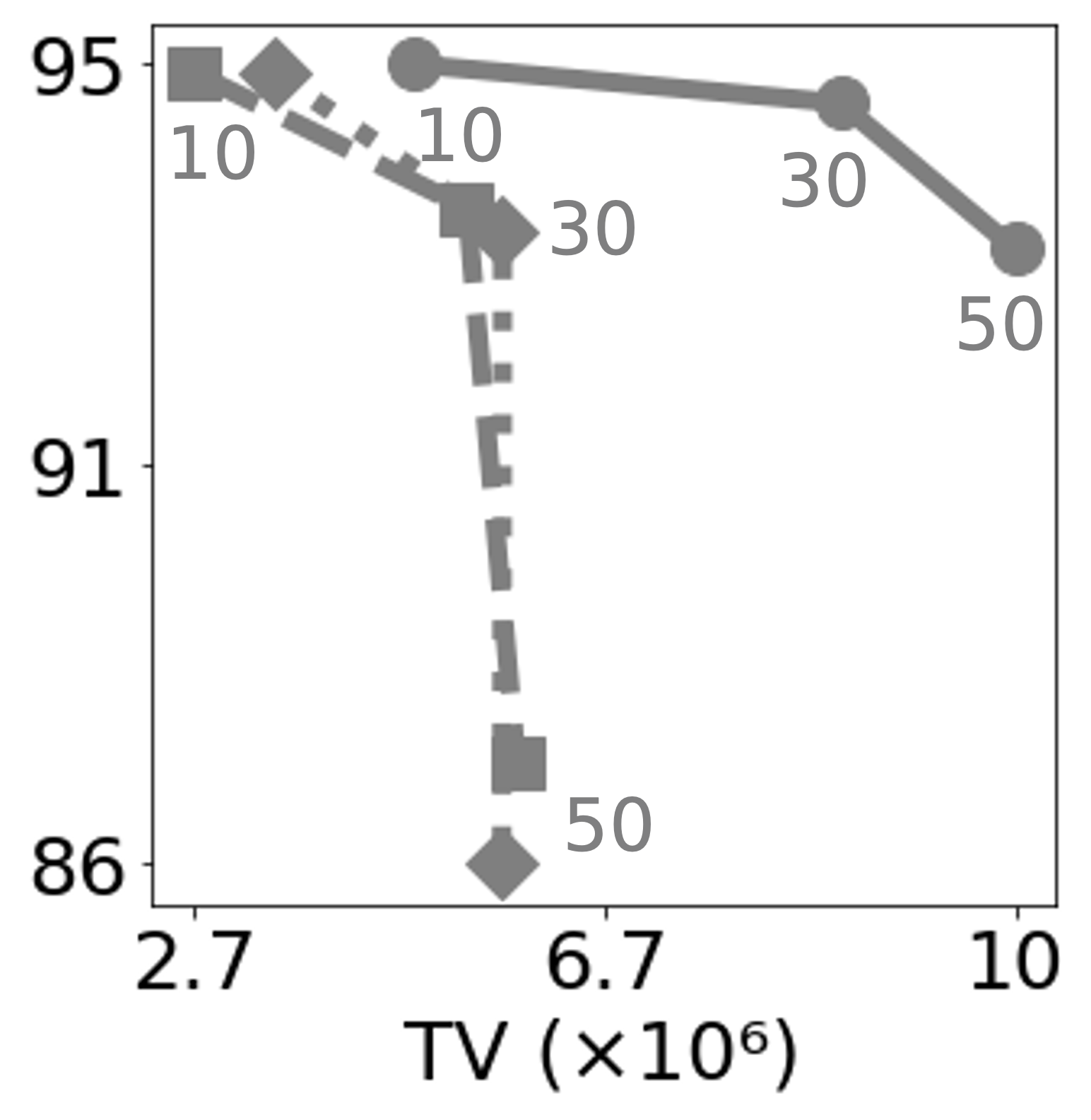}} &
\subfigure[Rand]{\includegraphics[width=0.30\columnwidth]{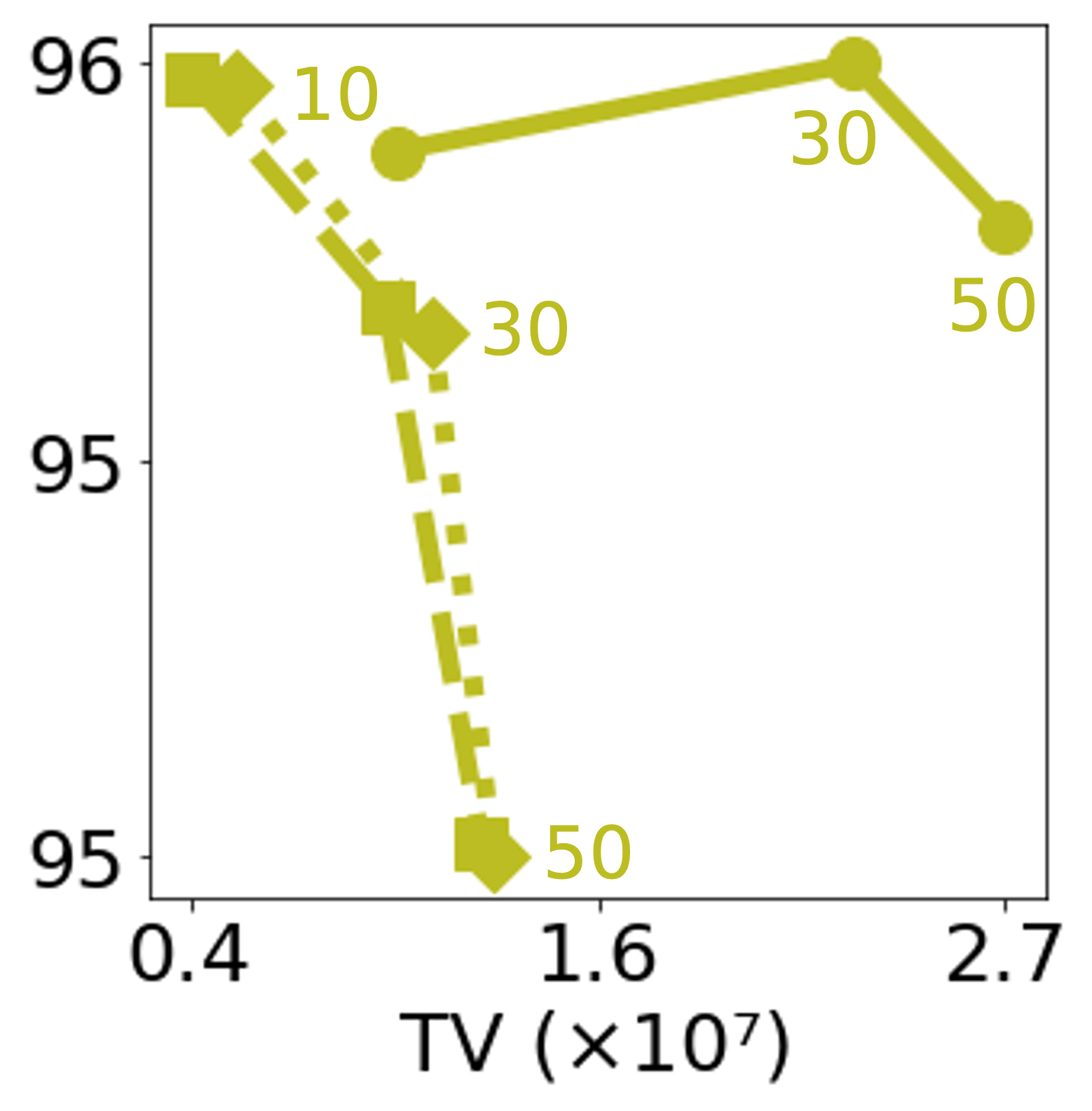}}
\end{tabular}\\
\subfigure{\includegraphics[width=0.7\columnwidth]{figure/1/legend.png}}
\caption{Effects of Gaussian filtering and max-pooling on total variation of attribution masks and model accuracy on SVHN.}
\label{fig:tv-accuracy-svhn}
\end{figure}
\vfil}
\newpage

\begin{figure*}[ht!]
\centering
\begin{tabular}{c | c  c  c}\hline
 & \small \bf 10\% & \small \bf 30\% & \small \bf 50\% \\\hline
\small \bf \rotatebox[origin=l]{90}{CIFAR-10} & \subfigure[\small $R^2=0.84$]{\includegraphics[width=0.20\textwidth]{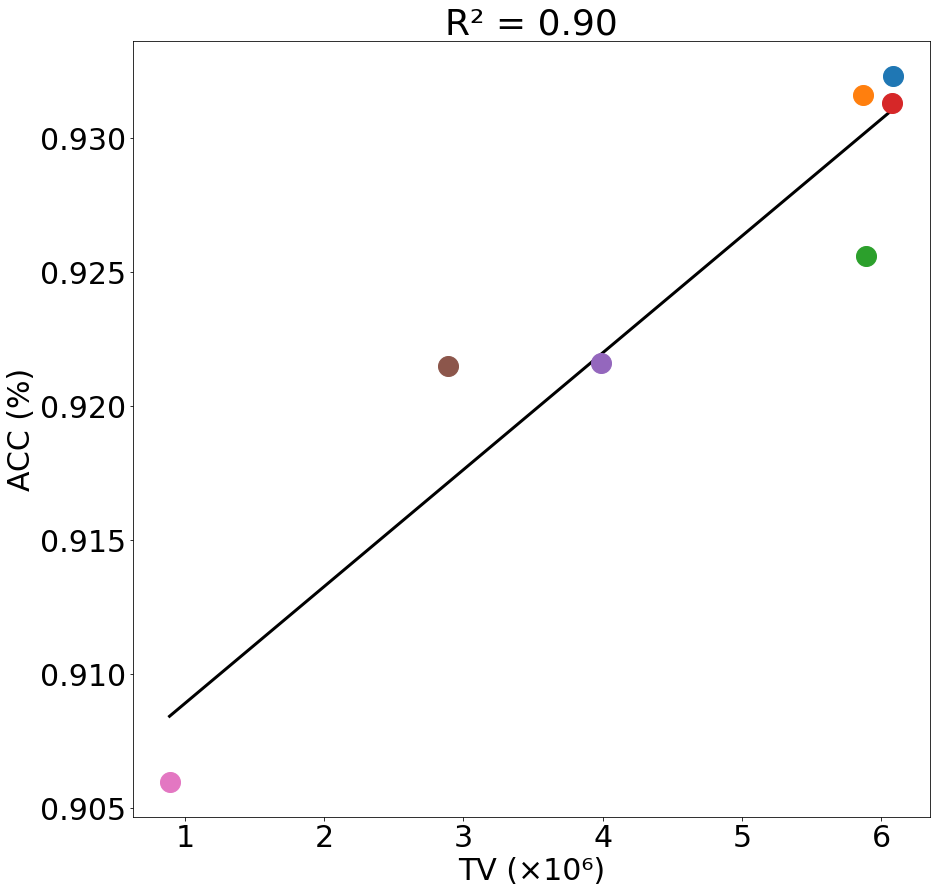}} &
\subfigure[$R^2=0.85$]{\includegraphics[width=0.20\textwidth]{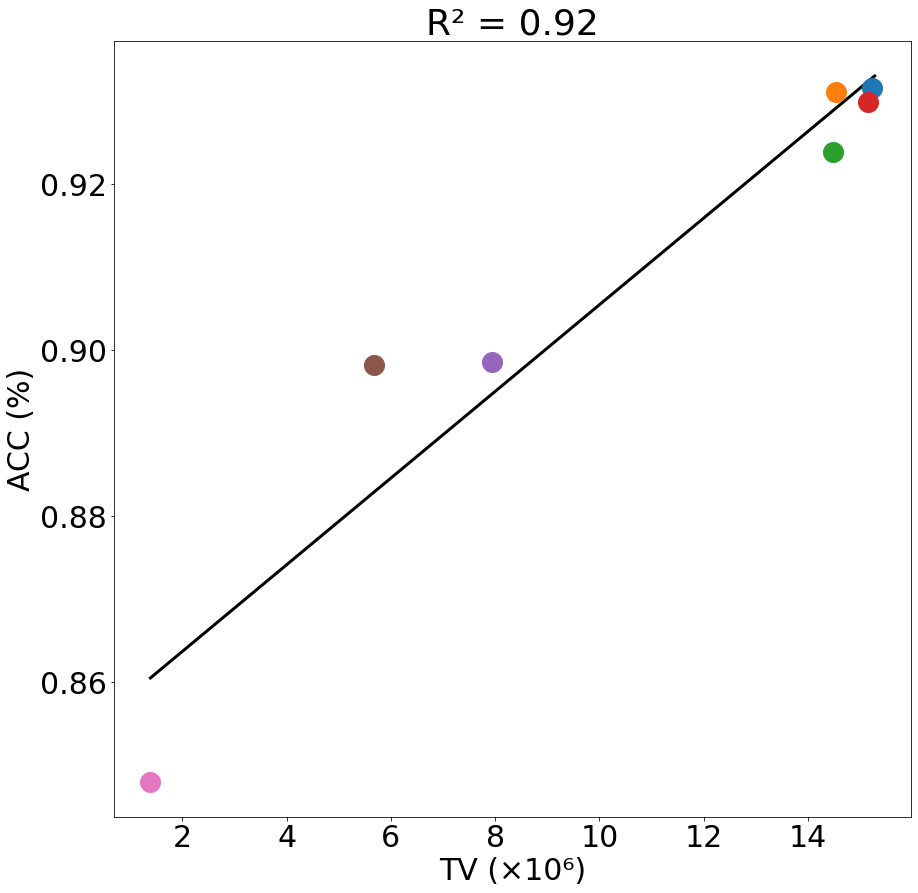}} &
\subfigure[$R^2=0.82$]{\includegraphics[width=0.20\textwidth]{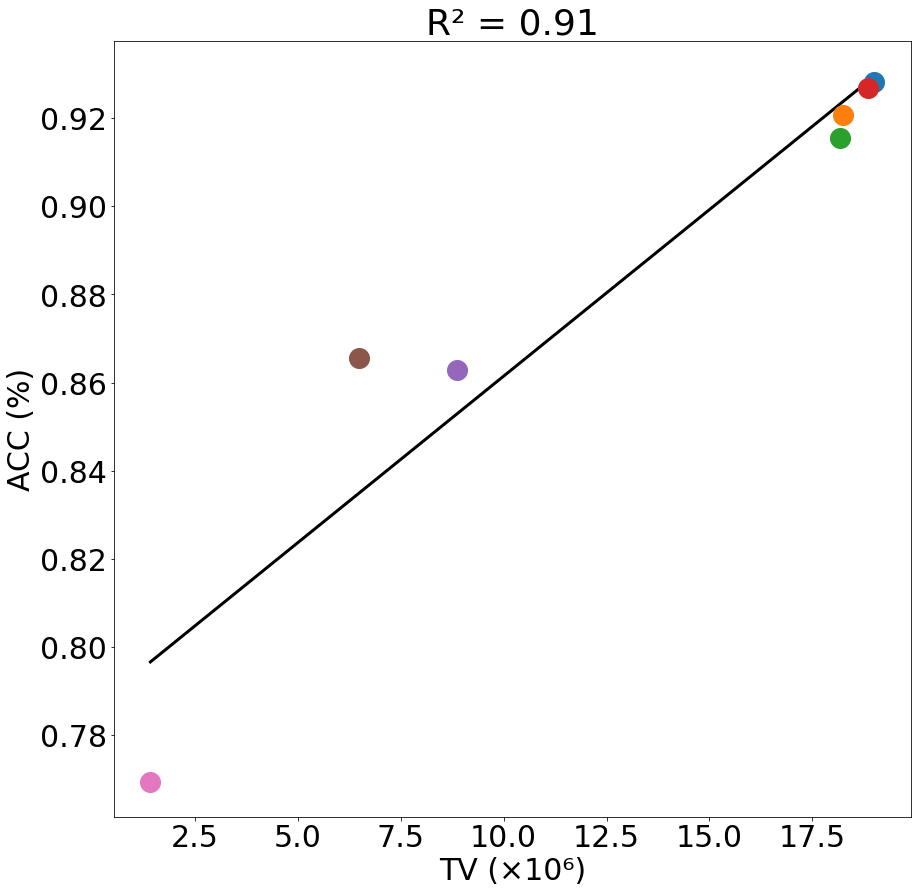}}\\
\small \bf \rotatebox[origin=l]{90}{SVHN} & \subfigure[$R^2=0.92$]{\includegraphics[width=0.20\textwidth]{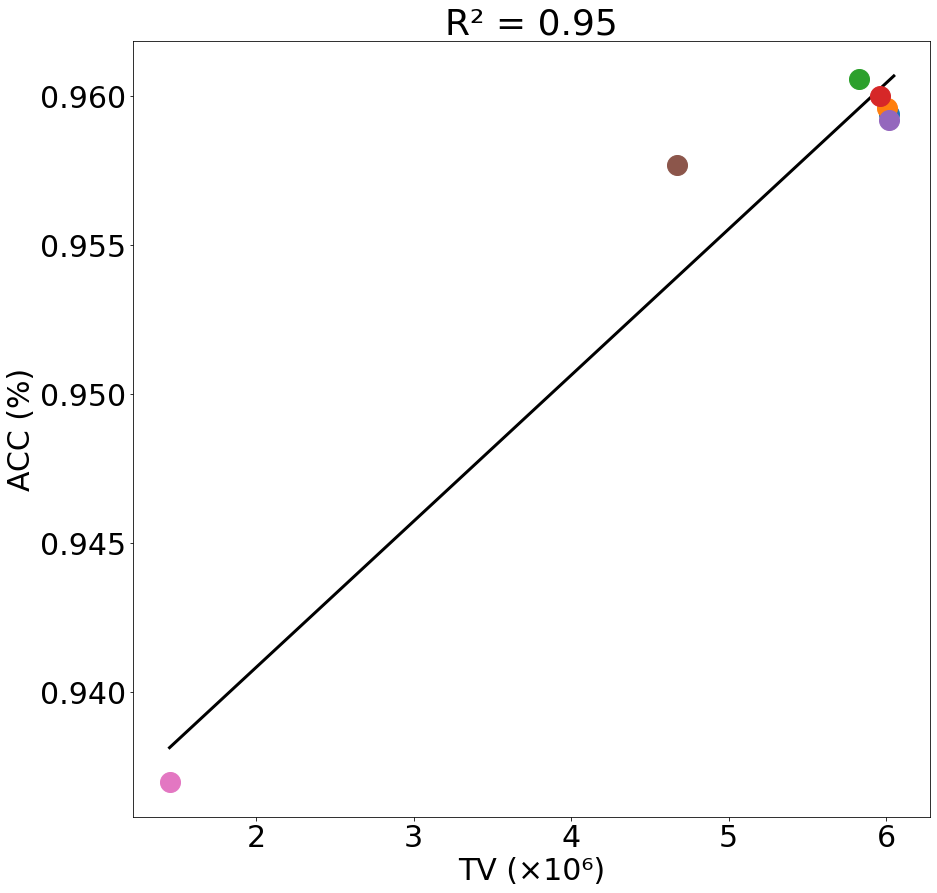}} &
\subfigure[$R^2=0.90$]{\includegraphics[width=0.20\textwidth]{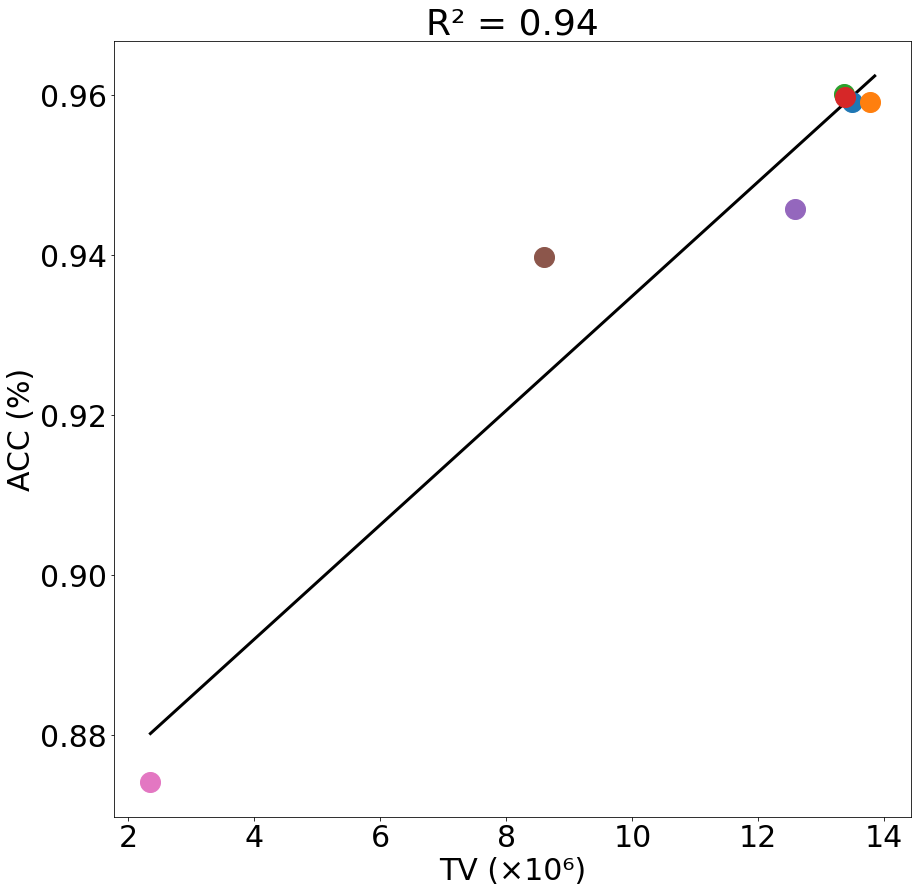}} &
\subfigure[$R^2=0.95$]{\includegraphics[width=0.20\textwidth]{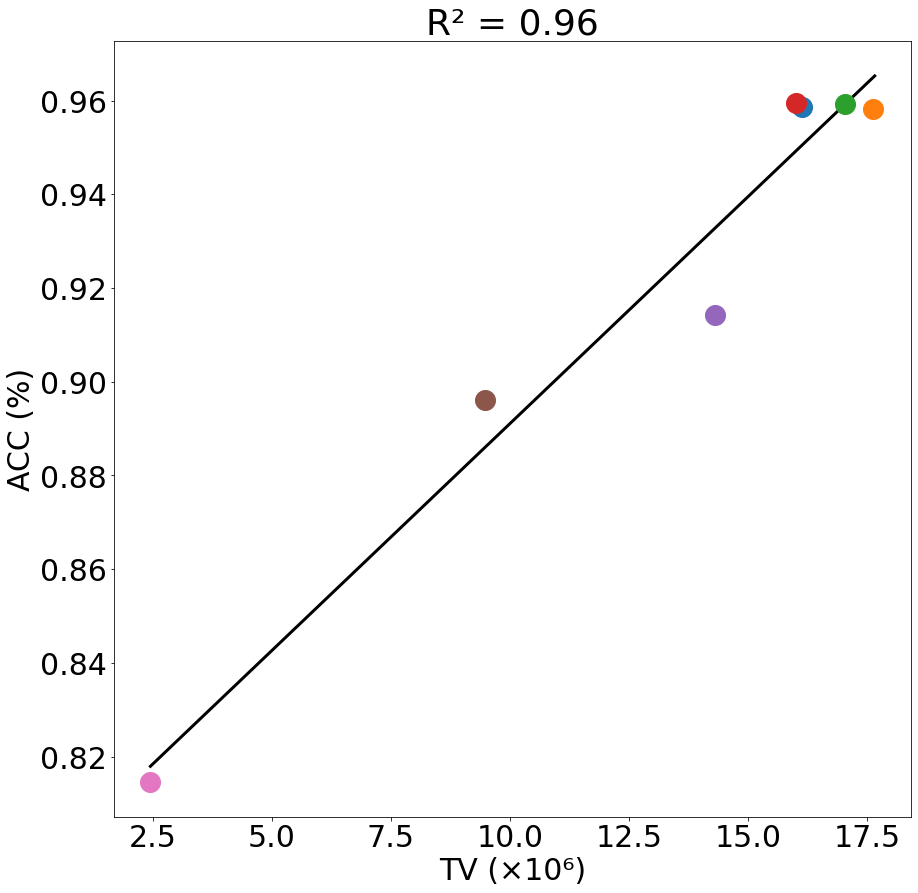}}\\
\small \bf \rotatebox[origin=l]{90}{CUB-200} & \subfigure[$R^2=0.91$]{\includegraphics[width=0.20\textwidth]{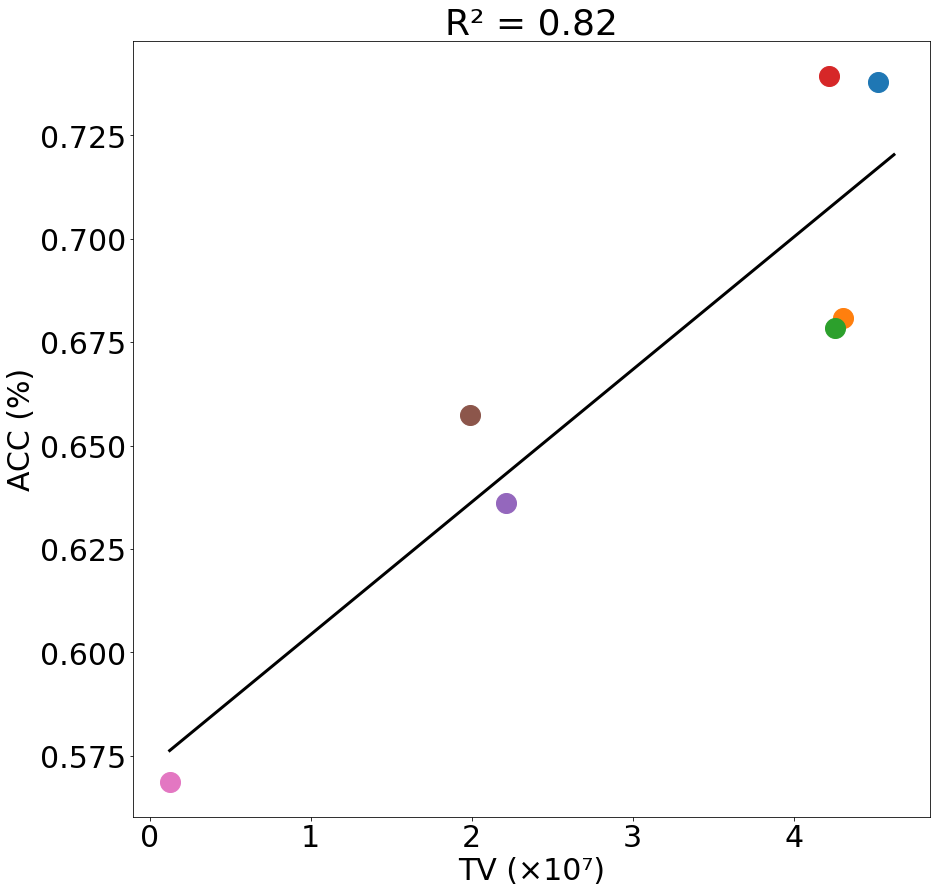}} &
\subfigure[$R^2=0.90$]{\includegraphics[width=0.20\textwidth]{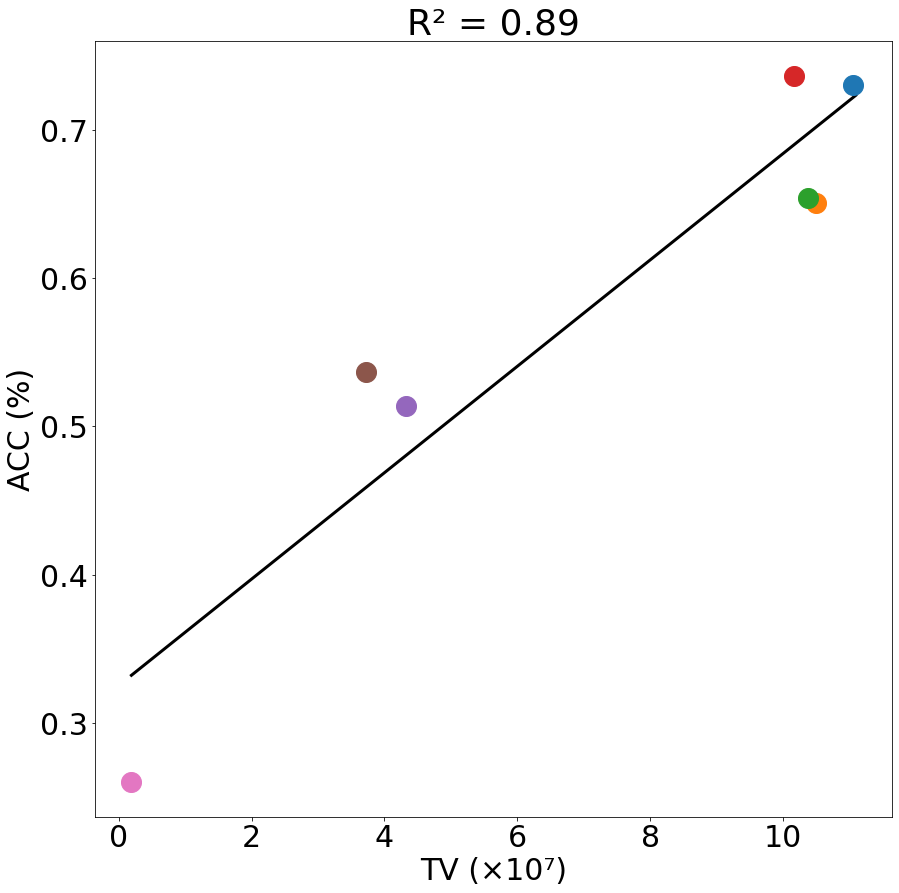}} &
\subfigure[$R^2=0.92$]{\includegraphics[width=0.20\textwidth]{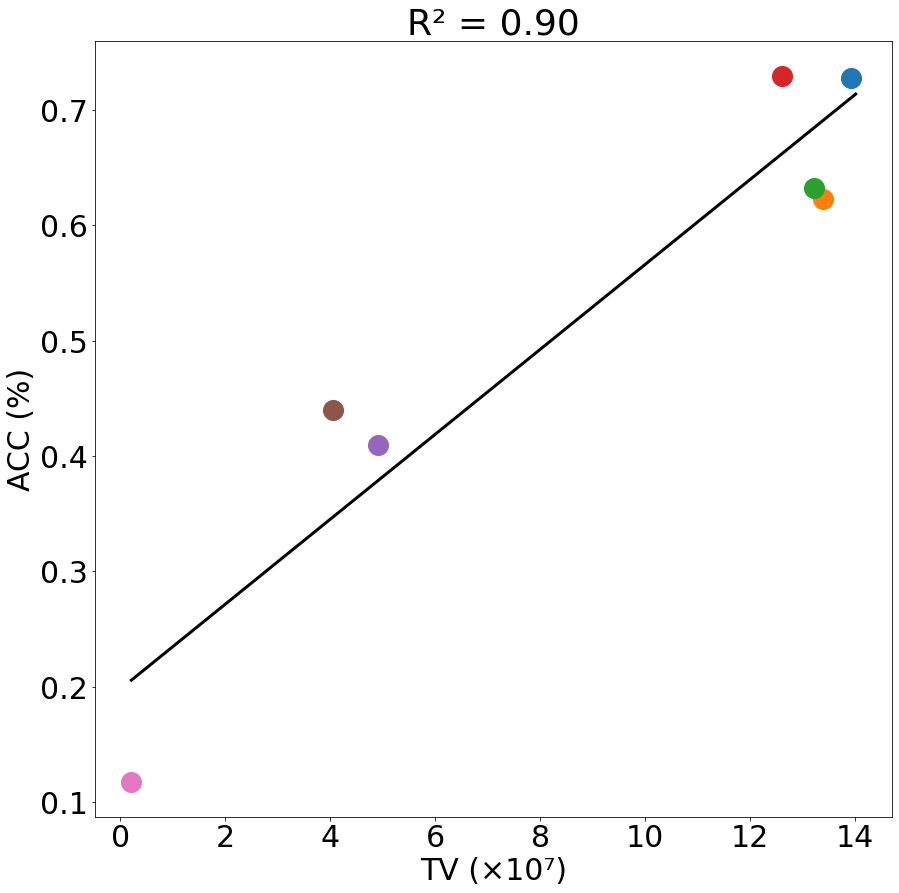}}
\end{tabular}\\
\subfigure{\includegraphics[width=0.55\textwidth]{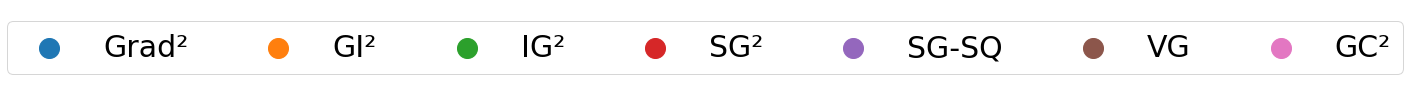}}
\caption{Relationship between model accuracy and total variation of attribution masks (\textit{TV}) in the aspect of non-squaring attribution method. The fitted line and coefficient of determination from simple linear regression are presented together. The y-axis represents the final test accuracy (\%) in the ROAR evaluation protocol.}
\label{fig:relation}
\end{figure*}

\end{document}